\documentclass[10pt, a4paper]{article}

\usepackage[]{lrec-coling2024} %

\usepackage{microtype}
\usepackage{inconsolata}

\usepackage{booktabs}
\usepackage{amsmath}
\usepackage{amssymb}
\usepackage{siunitx}
\usepackage{graphicx}
\usepackage{colortbl}
\usepackage{color}
\usepackage{tabularray}
\usepackage{tabularx}
\usepackage{multirow}
\usepackage{hhline}
\usepackage{makecell}
\usepackage{adjustbox}
\usepackage{enumitem}
\usepackage{subcaption}
\usepackage{float}
\usepackage{dblfloatfix}
\usepackage{authblk}
\UseTblrLibrary{booktabs}
\usepackage[font=small,skip=2pt]{caption}
\PassOptionsToPackage{hyphens}{url}\usepackage{hyperref}
\usepackage{enumitem}

\usepackage[noabbrev,nameinlink,capitalise]{cleveref}

\definecolor{Gray}{rgb}{0.501,0.501,0.501}
\definecolor{Rajah}{rgb}{0.968,0.737,0.415}
\definecolor{Apricot}{rgb}{0.913,0.541,0.443}
\definecolor{Goldenrod}{rgb}{0.996,0.823,0.403}
\definecolor{Sunglo}{rgb}{0.901,0.486,0.45}
\definecolor{SilverTree}{rgb}{0.341,0.733,0.541}
\definecolor{RobRoy}{rgb}{0.909,0.827,0.419}

\definecolor{rlt42}{HTML}{E67D74}
\definecolor{r42_48}{HTML}{E88A82}
\definecolor{r48_54}{HTML}{EFACA6}
\definecolor{r54_60}{HTML}{F8DBD9}
\definecolor{w60_63}{HTML}{F8F8F8}
\definecolor{g63_66}{HTML}{F1FAF6}
\definecolor{g66_69}{HTML}{DFF3E8}
\definecolor{g69_72}{HTML}{BFE6D3}
\definecolor{g72_75}{HTML}{A0D9BD}
\definecolor{g75_78}{HTML}{78CEA4}
\definecolor{rgt78}{HTML}{57BB8A}

\title{{\includegraphics[scale=0.08]{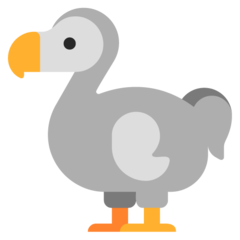}}\textbf{\textsc{DoDo} Learning: \textsc{Do}main-\textsc{D}em\textsc{o}graphic Transfer in Language Models for Detecting Abuse Targeted at Public Figures}}

\name{Angus R. Williams$^{1}$, Hannah Rose Kirk$^{1,2}$, Liam Burke-Moore$^{1}$, Yi-Ling Chung$^{1}$, \\{\bf \large Ivan Debono$^{1,3}$}, {\bf \large Pica Johansson$^{1}$}, {\bf \large Francesca Stevens$^{1}$}, {\bf \large Jonathan Bright$^{1}$}, \\{\bf \large Scott A. Hale$^{1,2}$}}

\address{$^{1}$The Alan Turing Institute, London, UK \\
         $^{2}$Oxford Internet Institute, University of Oxford, Oxford, UK \\
         $^{3}$Ofcom*, London, UK (work done while seconded to The Alan Turing Institute)\\
         angusrwilliams@gmail.com, onlinesafety@turing.ac.uk}

\abstract{
Public figures receive disproportionate levels of abuse on social media, impacting their active participation in public life. Automated systems can identify abuse at scale but labelling training data is expensive and potentially harmful. So, it is desirable that systems are efficient and generalisable, handling shared and specific aspects of abuse. We explore the dynamics of cross-group text classification in order to understand how well models trained on one domain or demographic can transfer to others, with a view to building more generalisable abuse classifiers. We fine-tune language models to classify tweets targeted at public figures using our novel \textsc{DoDo} dataset, containing $28{,}000$ entries with fine-grained labels, split equally across four \textsc{Do}main-\textsc{D}em\textsc{o}graphic pairs (male and female footballers and politicians). We find that (i) small amounts of diverse data are hugely beneficial to generalisation and adaptation; (ii) models transfer more easily across demographics but cross-domain models are more generalisable; (iii) some groups contribute more to generalisability than others; and (iv) dataset similarity is a signal of transferability.
 \\ \newline \Keywords{cross-domain, abuse detection, generalisability}}

\begin{document}

\maketitleabstract

\def\thefootnote{*}\footnotetext{The views and opinions in this paper are those of the author. They do not necessarily represent those of Ofcom, and are not statements of Ofcom policy.}\def\thefootnote{\arabic{footnote}}

{\footnotesize
\noindent\textcolor{red}{\textbf{Content Warning:} We include some synthetic examples of the dataset schema to illustrate its contents.}

\noindent\textbf{Data Release Statement:} Due to institutional guidelines concerning privacy issues (\cref{section:datarelease}), we are unable to release the \textsc{DoDo} dataset.
}

\section{Introduction}
Civil discussion between public figures and citizens is a key component of a well-functioning democratic society \cite{deweyPublic1927, roweCivility2015, papacharissi2004democracy}. Social media has opened new channels of communication and permitted greater access between users and public figures \cite{doidgeIf2015, wardTurds2020}; becoming an important tool for self-promotion, message spreading and maintaining a dialogue with fans, followers or the electorate \cite{farringtonSport2014}, beyond traditional media gatekeeping \cite{colemanCan1999, colemanNew2005, colemanExploring2003,williamsonEffect2009}. However, there is a cost: the immediacy, ease and anonymity of online interactions has routinised the problem of abuse \cite{sulerOnline2004, shulmanCase2009, brownWWW2009, joinsonOxford2012, roweCivility2015, wardTurds2020}. Public figures attract more intrusive and abusive attention than average users of online platforms \cite{mullenFixated2009, meloyStalking2008}, and abuse directed towards them is both highly-public yet often grounded in highly-personal attacks \cite{eriksonThree2021}. There are detrimental effects to individual victims' mental health, which can ultimately result in their withdrawal from public life \cite{vidgenIntroducing2021, delisleLargescale2019}, and to society from normalising a culture of abuse and hate \cite{ingleSports2021}. Disengagement is particularly worrisome for the functioning of democracy and political representation as it might be spread unevenly across groups \cite{theocharisBad2016, greenwoodOnline2019, wardTurds2020}, e.g. women MPs being more likely to leave politics than men \cite{manningMPs2019}.

Tackling abuse against public figures is a pressing issue, but the volume of social media posts makes manual investigations challenging, and conclusions drawn from anecdotal self-reporting or small sample size surveys offer limited and potentially biased coverage of the problem \cite{wardTurds2020}. Automated systems based on machine learning or language models can be used to classify text at scale, but depend on labelling training data which is complex, expensive to collect and potentially psychologically harmful to annotators \cite{kirkMoreDataBetter2022}.

In this context, it is highly desirable to develop abuse classifiers that can perform well across a range of different target groups whilst being trained on a minimal `labelling budget'. However, this may be technically challenging because, while some properties of abuse are shared across settings, different \textit{domains} (e.g., sport, politics or journalism) introduce linguistic and distributional shifts. Furthermore, previous reports reveal that the nature of online abuse is heavily influenced by the identity attributes of its targets, for example gendered abuse against female politicians \cite{bardallGenderSpecific2013, stambolievaMethodology2017, eriksonThree2021, delisleLargescale2019}; so, learnings from different \textit{demographics} may also not transfer. Exploring the effect of distributional shifts on model performance is useful for computational social scientists studying real-world phenomena, and for policymakers aiming to understand how to tackle online harm.

Despite the promise of generalisable abuse models for protecting more groups from harm, existing research focuses on fuzzy, keyword based definitions of domains, leading to datasets sourced around topics as opposed to target groups, and there is a lack of systematic study on the extent to which models trained on some combination of target groups can transfer to others. In this paper, we ask how well classifiers trained on data from specific factorisations of groups of public figures can transfer to others, with a view to building more generalisable models. Our novel \textsc{DoDo} dataset is collected from Twitter/X\footnote{Twitter has recently rebranded as "X". As the DoDo dataset was collected before the rebrand, we refer to the platform as Twitter exclusively.} and contains tweets targeted at public figures across two \textsc{Do}mains (UK members of parliament or ``MPs'', and professional footballer players) and two \textsc{D}em\textsc{o}graphic groups (women and men). Tweets are annotated with four fine-grained labels to disambiguate abuse from other sentiments like criticism.
We present results from experiments exploring the impacts of data diversity and number of training examples on domain-demographic transfer and generalisability.

\section{Dataset}

\subsection{Data Collection}
Our data is collected from Twitter. While generally over-researched \cite{vidgenDirections2020}, it is a dominant source for interactions between public figures and the general public. Most MPs have Twitter accounts and Twitter activity may even have a small impact on elections \cite{bright2020campaigning}. 

We compiled lists of accounts for UK MPs (590 accounts, 384 men, 206 women) and for players from England's top football divisions (808 from the Men's Premier League, 216 from the Women's Super League). We used the Twitter API Filtered Stream and Full Archive Search endpoints to collect all tweets that mention a public figure's account over a given time window.\footnote{A similar approach is adopted in prior work that tracks public figure abuse \cite{gorrellMP2020, wardTurds2020, rheaultPoliticians2019}.} 

Levels of abusive content `in-the-wild' are relatively low \citep{vidgen2019challenges}. In order to evaluate classifiers on realistic distributions while maximising their ability to detect abusive content, we randomly sample the test and validation datasets (preserving real-world class imbalance) but apply boosted sampling for the training dataset (ensuring the model sees enough instances of the rarer abusive class). We sample $7,000$ tweets in total for each domain-demographic pair: a $3,000$ train split, a $3,000$ test split, and a $1,000$ validation split.

\cref{section:add_dataset} provides more detail on data collection, processing, and sampling.

\subsection{Data Annotation}
\label{section:schema}

In the context of abuse detection, fine-grained labels can provide clarity for annotators, and enable more extensive error analysis, compared to binary labels. We employed annotators to label tweets with one of 4 classes of sentiment expressed towards public figures: Positive, Neutral, Critical, or Abusive, as defined below.\footnote{Labels are assigned based on the use of language, not the target of sentiment expressed.} 

\begin{enumerate}[nolistsep, labelindent=2pt, leftmargin=12pt]
\item \textbf{Positive:} Language that expresses support, praise, respect or encouragement towards an individual or group. It can praise specific skills, behaviours, or achievements, as well as encourage diversity and the representation of identities.

\item \textbf{Neutral:} Language with an unemotive tone or that does not fit the criteria of the other three categories, including factual statements, event descriptions, questions or objective remarks.

\item \textbf{Critical:} Language that makes a substantive negative assessment or claim about an individual or group. Negative assessment can be based on factors such as behaviour, performance, responsibilities, or actions, without being abusive.\footnote{The annotator guidelines focused on distinguishing between abuse and criticism. Criticism must include a rationale for negative opinions on an individual's actions (not their identity)---it is not a form of ``soft'' abuse.} %
\item \textbf{Abusive:} Language containing threats, insults, derogatory remarks (e.g., hateful use of slurs and negative stereotypes), dehumanisation (e.g., comparing individuals to insects, animals, or trash), mockery, or belittlement towards an individual, group, or protected identity attribute (The Equality Act (\citeyear{EqualityAct2010})). \newline
\end{enumerate}

\noindent The two domains were annotated sequentially in batches, but we updated our approach after the first batch as we found that crowdworkers struggled with the complexity of our task (see \cref{section:add_annotation} for details). The final Cohen Kappa\footnote{Calculated using the generalised formula from \citet{gwet_handbook_2014} to account for variable \# of annotations per entry.} for each domain was 0.50 for footballers and 0.67 for MPs.

\subsection{Analysis}

\paragraph{Terminology} We abbreviate pairs of domain-demographic data as: fb-m (footballers-men), fb-w (footballers-women), mp-m (MPs-men), mp-w (MPs-women). We refer to any given domain-demographic pair as a \texttt{dodo}. We refer to groups of models that we train by the number of \texttt{dodos} included in the training data: \texttt{dodo1} for models trained using one domain-demographic pair, \texttt{dodo2} for models trained using two pairs, etc.

\paragraph{Overview} The total dataset has 28,000 annotated entries, 7,000 for each \texttt{dodo} pair, with 3K/3K/1K test/train/validation splits. \cref{table:tweet_counts} shows class distributions across splits and counts of tweets sampled randomly pre-annotation.

\begin{table}[!t]
\centering
\setlength{\tabcolsep}{2.5pt}
\begin{adjustbox}{width=.48\textwidth}
\arrayrulecolor{black}
\begin{tabular}{llrr|rr|rr|rr} 
\hline
\multirow{2}{*}{\textbf{ Split}} & \multirow{2}{*}{\textbf{ Stance}} & \multicolumn{8}{c}{dodo}                                                                                                                                                                                                                   \\ 
\arrayrulecolor[rgb]{0.502,0.502,0.502}\cline{3-10}
                                 &                                   & \multicolumn{2}{c}{\textit{fb-m }}                      & \multicolumn{2}{c}{\textit{fb-w }}                      & \multicolumn{2}{c}{\textit{mp-m }}                      & \multicolumn{2}{c}{\textit{mp-w }}                       \\ 
\arrayrulecolor{black}\hline
\multirow{4}{*}{Train}           & Abusive                           & 867  & \textcolor[rgb]{0.502,0.502,0.502}{\textit{29\%}} & 481  & \textcolor[rgb]{0.502,0.502,0.502}{\textit{16\%}} & 1007 & \textcolor[rgb]{0.502,0.502,0.502}{\textit{34\%}} & 870  & \textcolor[rgb]{0.502,0.502,0.502}{\textit{29\%}}  \\
                                 & Critical                          & 475  & \textcolor[rgb]{0.502,0.502,0.502}{\textit{16\%}} & 282  & \textcolor[rgb]{0.502,0.502,0.502}{\textit{9\%}}  & 1283 & \textcolor[rgb]{0.502,0.502,0.502}{\textit{43\%}} & 1353 & \textcolor[rgb]{0.502,0.502,0.502}{\textit{45\%}}  \\
                                 & Neutral                           & 647  & \textcolor[rgb]{0.502,0.502,0.502}{\textit{21\%}} & 719  & \textcolor[rgb]{0.502,0.502,0.502}{\textit{24\%}} & 605  & \textcolor[rgb]{0.502,0.502,0.502}{\textit{20\%}} & 628  & \textcolor[rgb]{0.502,0.502,0.502}{\textit{21\%}}  \\
                                 & Positive                          & 1011 & \textcolor[rgb]{0.502,0.502,0.502}{\textit{34\%}} & 1518 & \textcolor[rgb]{0.502,0.502,0.502}{\textit{51\%}} & 105  & \textcolor[rgb]{0.502,0.502,0.502}{\textit{3\%}}  & 149  & \textcolor[rgb]{0.502,0.502,0.502}{\textit{5\%}}   \\ 
\hline
\multirow{4}{*}{Test}            & Abusive                           & 103  & \textcolor[rgb]{0.502,0.502,0.502}{\textit{3\%}}  & 43   & \textcolor[rgb]{0.502,0.502,0.502}{\textit{1\%}}  & 392  & \textcolor[rgb]{0.502,0.502,0.502}{\textit{13\%}} & 373  & \textcolor[rgb]{0.502,0.502,0.502}{\textit{12\%}}  \\
                                 & Critical                          & 377  & \textcolor[rgb]{0.502,0.502,0.502}{\textit{13\%}} & 89   & \textcolor[rgb]{0.502,0.502,0.502}{\textit{3\%}}  & 1467 & \textcolor[rgb]{0.502,0.502,0.502}{\textit{49\%}} & 1471 & \textcolor[rgb]{0.502,0.502,0.502}{\textit{49\%}}  \\
                                 & Neutral                           & 811  & \textcolor[rgb]{0.502,0.502,0.502}{\textit{27\%}} & 767  & \textcolor[rgb]{0.502,0.502,0.502}{\textit{26\%}} & 985  & \textcolor[rgb]{0.502,0.502,0.502}{\textit{33\%}} & 927  & \textcolor[rgb]{0.502,0.502,0.502}{\textit{31\%}}  \\
                                 & Positive                          & 1709 & \textcolor[rgb]{0.502,0.502,0.502}{\textit{57\%}} & 2101 & \textcolor[rgb]{0.502,0.502,0.502}{\textit{70\%}} & 156  & \textcolor[rgb]{0.502,0.502,0.502}{\textit{5\%}}  & 229  & \textcolor[rgb]{0.502,0.502,0.502}{\textit{8\%}}   \\ 
\hline
\multirow{4}{*}{Validation}      & Abusive                           & 33   & \textcolor[rgb]{0.502,0.502,0.502}{\textit{3\%}}  & 14   & \textcolor[rgb]{0.502,0.502,0.502}{\textit{1\%}}  & 140  & \textcolor[rgb]{0.502,0.502,0.502}{\textit{14\%}} & 135  & \textcolor[rgb]{0.502,0.502,0.502}{\textit{13\%}}  \\
                                 & Critical                          & 93   & \textcolor[rgb]{0.502,0.502,0.502}{\textit{9\%}}  & 45   & \textcolor[rgb]{0.502,0.502,0.502}{\textit{5\%}}  & 484  & \textcolor[rgb]{0.502,0.502,0.502}{\textit{48\%}} & 459  & \textcolor[rgb]{0.502,0.502,0.502}{\textit{46\%}}  \\
                                 & Neutral                           & 335  & \textcolor[rgb]{0.502,0.502,0.502}{\textit{34\%}} & 267  & \textcolor[rgb]{0.502,0.502,0.502}{\textit{27\%}} & 332  & \textcolor[rgb]{0.502,0.502,0.502}{\textit{33\%}} & 337  & \textcolor[rgb]{0.502,0.502,0.502}{\textit{34\%}}  \\
                                 & Positive                          & 539  & \textcolor[rgb]{0.502,0.502,0.502}{\textit{54\%}} & 674  & \textcolor[rgb]{0.502,0.502,0.502}{\textit{67\%}} & 44   & \textcolor[rgb]{0.502,0.502,0.502}{\textit{4\%}}  & 69   & \textcolor[rgb]{0.502,0.502,0.502}{\textit{7\%}}   \\ 
\hline\hline
\multirow{4}{*}{Random}          & Abusive                           & 181  & \textcolor[rgb]{0.502,0.502,0.502}{\textit{3\%}}  & 75   & \textcolor[rgb]{0.502,0.502,0.502}{\textit{1\%}}  & 744  & \textcolor[rgb]{0.502,0.502,0.502}{\textit{13\%}} & 661  & \textcolor[rgb]{0.502,0.502,0.502}{\textit{12\%}}  \\
                                 & Critical                          & 642  & \textcolor[rgb]{0.502,0.502,0.502}{\textit{12\%}} & 197  & \textcolor[rgb]{0.502,0.502,0.502}{\textit{4\%}}  & 2676 & \textcolor[rgb]{0.502,0.502,0.502}{\textit{49\%}} & 2676 & \textcolor[rgb]{0.502,0.502,0.502}{\textit{49\%}}  \\
                                 & Neutral                           & 1677 & \textcolor[rgb]{0.502,0.502,0.502}{\textit{30\%}} & 1466 & \textcolor[rgb]{0.502,0.502,0.502}{\textit{27\%}} & 1788 & \textcolor[rgb]{0.502,0.502,0.502}{\textit{33\%}} & 1741 & \textcolor[rgb]{0.502,0.502,0.502}{\textit{32\%}}  \\
                                 & Positive                          & 3000 & \textcolor[rgb]{0.502,0.502,0.502}{\textit{55\%}} & 3762 & \textcolor[rgb]{0.502,0.502,0.502}{\textit{68\%}} & 292  & \textcolor[rgb]{0.502,0.502,0.502}{\textit{5\%}}  & 422  & \textcolor[rgb]{0.502,0.502,0.502}{\textit{7\%}}   \\
\hline
\end{tabular}
\end{adjustbox}
\caption{Tweet counts across splits, \texttt{dodos}, and stances, with percentages within the \texttt{dodo} split. Includes counts and percentages for tweets from all splits selected by random sampling before annotation (5,500 tweets total per \texttt{dodo}).}
\label{table:tweet_counts}
\end{table}

\paragraph{Class Distributions}
\label{sec:classdist}
The last row of \cref{table:tweet_counts} contains the randomly sampled entries across each dataset (ignoring keyword sampled entries which would skew the distributions). The majority of tweets in the MPs datasets are abusive or critical, in contrast to the footballers datasets where the majority class is positive, especially for fb-w. We also see slightly higher proportions of abusive tweets targeted at male demographic groups (fb-m, mp-m). Further analysis here is outside the scope of this paper, but it is notable how levels of abuse vary.

\paragraph{Tweet Length}
The MPs data contains longer tweets on average than the footballers data (125 vs. 84 characters), and has over twice as many tweets $\ge$ 250 characters (1,632 vs. 556 tweets). 62\% of these longer ($\ge$250 characters) tweets for MPs are critical, implying the presence of detailed political debate.

\section{Experiments}
\label{section:experiments}

We conduct experiments to study how well model performance transfers across domains and demographics, and how the quantity and diversity of training data affects model generalisability across domains of public figures. To reflect the focus on generalisability, we evaluate models on: (i) ``seen'' \texttt{dodos} (test sets of \texttt{dodos} whose train sets were used in training); (ii) ``unseen'' \texttt{dodos} (test sets of \texttt{dodos} whose train sets were not used in training); and (iii) the total evaluation set (including test sets from all \texttt{dodos}). All test sets are fully held out from training---by ``seen'' and ``unseen'' we only mean the domain or demographic. We train models on data from combinations of \texttt{dodo} pairs, and experiment with continued fine-tuning on the resulting models. We repeat experiments across 3 random seeds and 2 labelling budgets. We make predictions using the total test set (12,000), and calculate mean and standard deviation of Macro-F1 across the seeds. The Macro-F1 score represents a macro-average of per class F1 scores, neutralising class imbalance. We also investigate the correlation of Macro-F1 with dataset similarity.

\paragraph{Models} We fine-tune deBERTa-v3 (\textbf{deBERT}, \citealp{he2021debertav3})\footnote{We also ran experiments on distilBERT \citep{sanh2019distilbert}, but deBERTa-v3 had consistently higher performance, therefore we only present results for deBERTa-v3.}, using Huggingface's Transformers Library\cite{huggingface_transformers}. We used Tesla K80 GPUs through Microsoft Azure, training for 5 epochs with an early stopping patience of 2 epochs using Macro-F1 on the validation set, requiring a total of 235 GPU hours.

\paragraph{Dodo Combinations} Our dataset has four \texttt{dodo} pairs, each with 3,000 training entries. There are 15 combinations of these pairs (if order does not matter): four single pairs (\texttt{dodo1}), six ways to pick two pairs (\texttt{dodo2}), four ways to pick three pairs (\texttt{dodo3}) and all pairs (\texttt{dodo4}). For all combinations, we randomly shuffle the concatenated training data before any training commences.

\paragraph{Labelling Budget} For each training combination, we make two budget assumptions. In the \textbf{full budget} condition, we concatenate the training sets: 3,000 training entries for \texttt{dodo1} experiments; 6,000 for \texttt{dodo2} experiments; 9,000 for \texttt{dodo3}; and 12,000 for \texttt{dodo4}. In the \textbf{fixed budget} condition, we assume train budget is fixed at $3{,}000$ entries and allocate ratios according to the \texttt{dodo} combinations: each included \texttt{dodo} makes up 100\% of the budget for \texttt{dodo1} experiments; 50\% for \texttt{dodo2}; 33\% for \texttt{dodo3}; and 25\% for \texttt{dodo4}. This allows us to test the effects of training data composition without confounding effects of its size.

\section{Results}
\label{section:results}

\subsection{Small amounts of diverse data are hugely beneficial to generalisable performance.}
\label{section:smalldiverse}

\cref{tab:focus_mf1} provides an overview of the performance of models trained on all combinations of \texttt{dodos}. The increase in performance from adding data from new domains or demographics is not linear: the full budget \texttt{dodo2} models only attain a one percentage point (pp) average increase in Macro-F1 Score for an additional 3,000 training entries. We also see the two \texttt{dodo4} models are only separated by 3pp despite the full budget version being exposed to 4 times the amount of training data as the fixed budget version. This shows that gains from data diversity outweigh those from significantly greater quantities of data in training generalisable models.

\begin{table}[!t]
\footnotesize
\centering
\begin{adjustbox}{width=\columnwidth}
\begin{tblr}{
  cells = {c},
  cell{1}{1} = {r=2}{},
  cell{1}{2} = {c=4}{},
  cell{1}{6} = {c=2}{},
  cell{3}{1} = {r=4}{},
  cell{7}{1} = {r=6}{},
  cell{13}{1} = {r=4}{},
  cell{3}{6} = {g66_69},
  cell{4}{6} = {w60_63},
  cell{5}{6} = {g63_66},
  cell{6}{6} = {g63_66},
  cell{7}{6} = {g66_69},
  cell{7}{7} = {g66_69},
  cell{8}{6} = {g66_69},
  cell{8}{7} = {g66_69},
  cell{9}{6} = {g72_75},
  cell{9}{7} = {g69_72},
  cell{10}{6} = {g69_72},
  cell{10}{7} = {g69_72},
  cell{11}{6} = {g72_75},
  cell{11}{7} = {g69_72},
  cell{12}{6} = {g69_72},
  cell{12}{7} = {g63_66},
  cell{13}{6} = {g69_72},
  cell{13}{7} = {g69_72},
  cell{14}{6} = {g72_75},
  cell{14}{7} = {g69_72},
  cell{15}{6} = {g72_75},
  cell{15}{7} = {g69_72},
  cell{16}{6} = {g72_75},
  cell{16}{7} = {g69_72},
  cell{17}{6} = {g72_75},
  cell{17}{7} = {g69_72},
  vline{2,6} = {1-18}{},
  vline{3-5} = {3-17}{Gray},
  hline{1,3,7,13,17,18} = {-}{},
  hline{4-6,8-12,14-16} = {2-8}{Gray},
}
{\textbf{Model}\\\textbf{Group}} & \textbf{Train on} &      &      &      & \textbf{Test On} &       \\
                     & \textit{fb-m} & \textit{fb-w} & \textit{mp-m} & \textit{mp-w} & \textit{Full} & \textit{Fixed}  \\
dodo1                & \checkmark                 &      &      &      & 0.676            & -     \\
                     &                   & \checkmark    &      &      & 0.612            & -     \\
                     &                   &      & \checkmark    &      & 0.655            & -     \\
                     &                   &      &      & \checkmark    & 0.643            & -     \\
dodo2                & \checkmark                 & \checkmark    &      &      & 0.667            & 0.673 \\
                     &                   &      & \checkmark    & \checkmark    & 0.675            & 0.661 \\
                     & \checkmark                 &      & \checkmark    &      & 0.723            & \textbf{0.708} \\
                     &                   & \checkmark    &      & \checkmark    & 0.718            & 0.698 \\
                     & \checkmark                 &      &      & \checkmark    & 0.722            & \textbf{0.708} \\
                     &                   & \checkmark    & \checkmark    &      & 0.718            & 0.654 \\
dodo3                & \checkmark                 & \checkmark    & \checkmark    &      & 0.702            & 0.695 \\
                     & \checkmark                 & \checkmark    &      & \checkmark    & 0.724            & 0.706 \\
                     & \checkmark                 &      & \checkmark    & \checkmark    & 0.727            & \textbf{0.708} \\
                     &                   & \checkmark    & \checkmark    & \checkmark    & 0.725            & 0.700 \\
dodo4                & \checkmark                 & \checkmark    & \checkmark    & \checkmark    & \textbf{0.731}            & 0.701 
\end{tblr}
\end{adjustbox}
\caption{\small Table of Macro-F1 scores on the total test set for all possible training data combinations, in both full and fixed budget scenarios. Colour-coded according to increasing Macro-F1 Score, with best scores for each budget in bold.}
\label{tab:focus_mf1}
\end{table}

\begin{table}[t]
\footnotesize
\centering
\begin{tblr}{
  row{1} = {c},
  row{2} = {c},
  column{3} = {c},
  column{5} = {c},
  cell{1}{1} = {r=2}{},
  cell{1}{2} = {c=4}{},
  cell{2}{2} = {c=2}{},
  cell{2}{4} = {c=2}{},
  cell{3}{3} = {g63_66},
  cell{3}{5} = {r54_60},
  cell{4}{3} = {g66_69},
  cell{4}{5} = {r48_54},
  cell{5}{3} = {g72_75},
  cell{5}{5} = {g72_75},
  cell{6}{3} = {g72_75},
  cell{6}{5} = {g69_72},
  hline{1,7} = {-}{0.08em},
  hline{2} = {2-5}{},
  hline{3,5} = {-}{},
}
\textbf{Train on}   & \textbf{Test on} &      &                  &      \\
                    & \textit{Seen}    &      & \textit{Unseen } &      \\
fb-m; fb-w\textbf{} & FBs      & 0.654 & MPs              & 0.576 \\
mp-m; mp-w          & MPs              & 0.682 & FBs      & 0.560 \\
fb-m; mp-m\textbf{} & Men              & 0.718 & Women            & 0.724 \\
fb-w; mp-w          & Women            & 0.722 & Men              & 0.690 
\end{tblr}
\caption{\small Cross-domain and cross-demographic transfer with mean Macro-F1 for full-budget \texttt{dodo2} models. We train on two \texttt{dodos} and evaluate on concatenated portions of the test set, e.g., we train \textit{fb-w; fb-m} then test on \textit{fb-w; fb-m} (seen) and \textit{mp-m, mp-w} (unseen). Colour-coded according to increasing Macro-F1 Score.\vspace{-1em}}
\label{tab:dodo2_compare}
\end{table}

\subsection{Cross-demographic transfer is more effective than cross-domain.}
\label{section:dodo2}
\cref{tab:dodo2_compare} shows the comparisons for domain transfer and demographic transfer by Macro-F1 score on the seen and unseen portions of the test set, using the full-budget \texttt{dodo2} models. For domain transfer, training on footballers gives a 0.654 F1 on the footballers dataset and 0.576 F1 on the MPs datasets. This is symmetric with training on MPs and testing on footballers. For demographic transfer, training on the male pairs and testing on female pairs faces no drop in performance. In contrast, training on women and testing on men leads to a small reduction in performance on the male data. In general, this demonstrates that transferring across domains is more challenging than transferring across demographics while keeping the domain fixed.

\begin{figure}[h]
    \centering
    \includegraphics[width = \columnwidth]{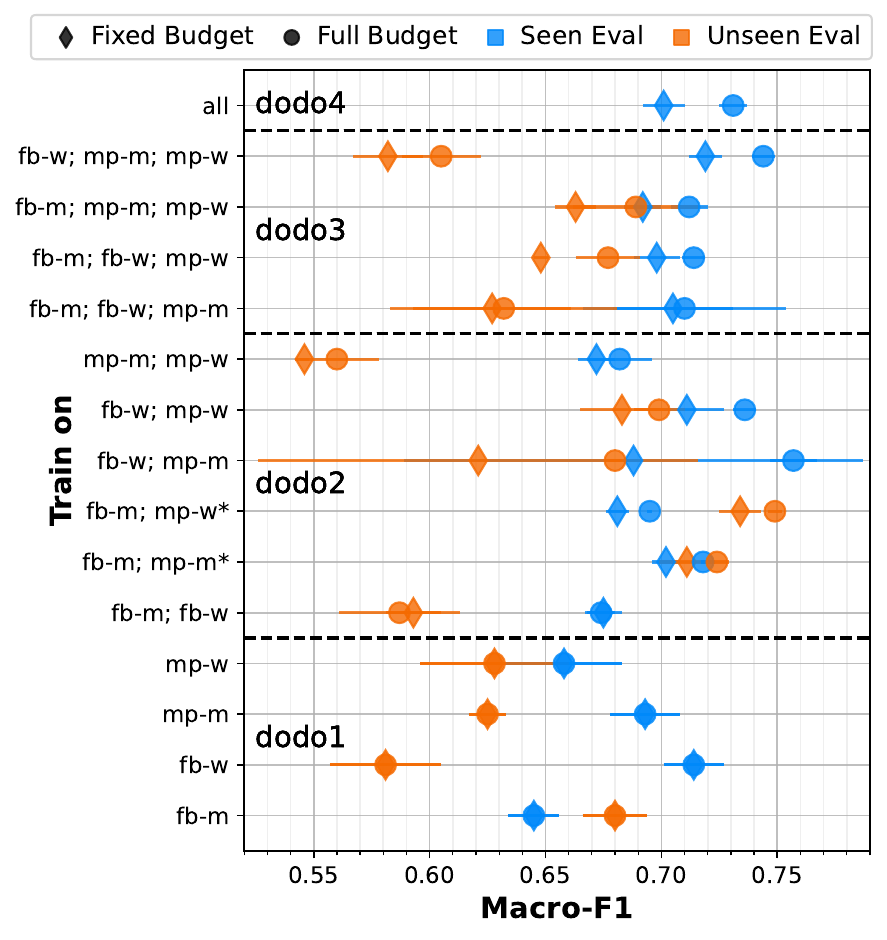}
    \caption{\small Mean and std-dev Macro-F1 across seeds for models trained on \texttt{dodo} combos, for fixed and full budgets, on test sets from seen and unseen \texttt{dodos}. *We removed one degenerate training seed (s=2).}
    \label{fig:dumbbells_single}
\end{figure}

\subsection{Cross-domain models are more generalisable than cross-demographic.}
\label{section:seenunseen}

\cref{fig:dumbbells_single} shows that, as expected, performance on test sets from seen \texttt{dodos} is generally higher than on those from unseen \texttt{dodos} (we investigate exceptions in \cref{section:add_unseen}). Within the \texttt{dodo2} models, cross-demographic within-domain models (e.g., \texttt{fb-m;fb-w}) perform 10pp better on average on seen \texttt{dodo} evaluation sets than unseen ones, compared to a much narrower gap of 1pp on average for cross-domain models (e.g., \texttt{fb-w;mp-w}). We also see from \cref{tab:focus_mf1} that cross-domain within-demographic \texttt{dodo2} models outperform all cross-demographic within-domain \texttt{dodo2} models on the total test set. This provides evidence that, within the context of this study, models trained on a single domain struggle to deal with out-of-domain examples, and that cross-domain models are more generalisable.

\subsection{Not all \texttt{dodos} contribute equally to generalisable performance.}
\label{section:notallequal}
\begin{figure}[h]
    \centering
    \includegraphics[width=\columnwidth]{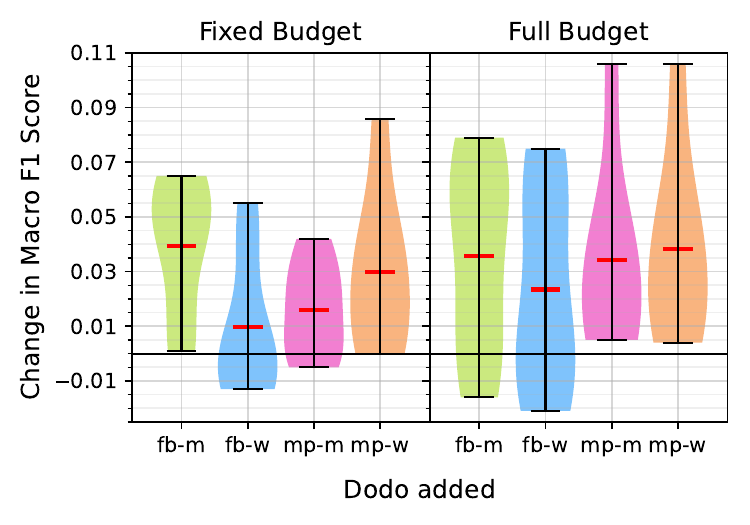}
    \caption{\small Violin plot displaying distribution of change in Macro-F1 score when adding a \texttt{dodo} to the training data (7 possible scenarios), with mean represented by red marker.}
    \label{fig:delta_violin}
\end{figure}

The average Macro-F1 increase provided by including each \texttt{dodo} in training is summarised in \cref{fig:delta_violin}. \texttt{fb-m} provides the largest average increase in a fixed budget scenario, and \texttt{mp-w} in a full budget scenario.\footnote{According to mean change in performance across all 7 possible scenarios of adding a \texttt{dodo} to training data.} In some cases, including fb-w data during training can detract from performance across both budgets. A \texttt{dodo1} model trained only on fb-m also outperforms all other \texttt{dodo1} models on the total test set (see \cref{tab:focus_mf1}), and fb-m data is included in the training dataset for the top ranking model for each dodo size across both labelling budgets. This suggests that training with fb-m is more important for good model generalisation than other \texttt{dodos}.

We now consider the situation of leaving out one \texttt{dodo} pair during training. We compare this left out case (\texttt{dodo3}) to training on all pairs (\texttt{dodo4}) in \cref{tab:leave_out}. We show the change in Macro-F1 on the total test set and change in number of training entries. For the full budget, leaving out mp-w from training leads to the largest reduction in performance. In contrast, removing all fb-w or mp-m entries does not significantly degrade performance even with 3,000 fewer training entries. For the fixed budget setting (with no confounding by training size), leaving out the two male pairs leads to a larger drop in performance than leaving out two female pairs.

\begin{table}[]
\centering
\footnotesize
\setlength{\extrarowheight}{0pt}
\addtolength{\extrarowheight}{\aboverulesep}
\addtolength{\extrarowheight}{\belowrulesep}
\setlength{\aboverulesep}{0pt}
\setlength{\belowrulesep}{0pt}
\begin{tabular}{lcccc} 
\toprule
                                            & \multicolumn{2}{c}{\textbf{Raw size}}                                             & \multicolumn{2}{c}{\textbf{Fixed size}}                                            \\ 
\cline{2-5}
                                            & \multicolumn{1}{l}{\textbf{$\Delta$ F1}}      & \multicolumn{1}{l}{\textbf{$\Delta$ N}} & \multicolumn{1}{l}{\textbf{$\Delta$ F1}}      & \multicolumn{1}{l}{\textbf{$\Delta$ N}}  \\ 
\hline \hline
\rowcolor[rgb]{0.753,0.753,0.753} all dodos & 0.731                                      & 12,000                                & 0.701                                      & 3,000                                  \\ 
\hline \hline
leave out fb-m                             & {\cellcolor[rgb]{0.933,0.655,0.631}}-0.006 & -3,000                                & {\cellcolor[rgb]{0.99,0.94,0.93}}-0.001    & 0                                     \\
leave out fb-w                             & {\cellcolor[rgb]{0.933,0.7,0.68}}-0.004    & -3,000                                & 0.007                                      & 0                                     \\
leave out mp-m                             & {\cellcolor[rgb]{0.933,0.62,0.58}}-0.007 & -3,000                                & 0.005                                      & 0                                     \\
leave out mp-w                             & {\cellcolor[rgb]{0.89,0.45,0.41}}-0.029 & -3,000                                & {\cellcolor[rgb]{0.933,0.655,0.631}}-0.006  & 0                                     \\
\bottomrule
\end{tabular}
\caption{\small Comparing  model trained on all pairs (\texttt{dodo4}) with models trained on 3 pairs (\texttt{dodo3}). Shows relative change in mean Macro-F1 on total test set, and relative change in $N$ of training entries.}
\label{tab:leave_out}
\end{table}

\begin{figure*}[t]
    \centering
    \includegraphics[width = 0.99\textwidth]{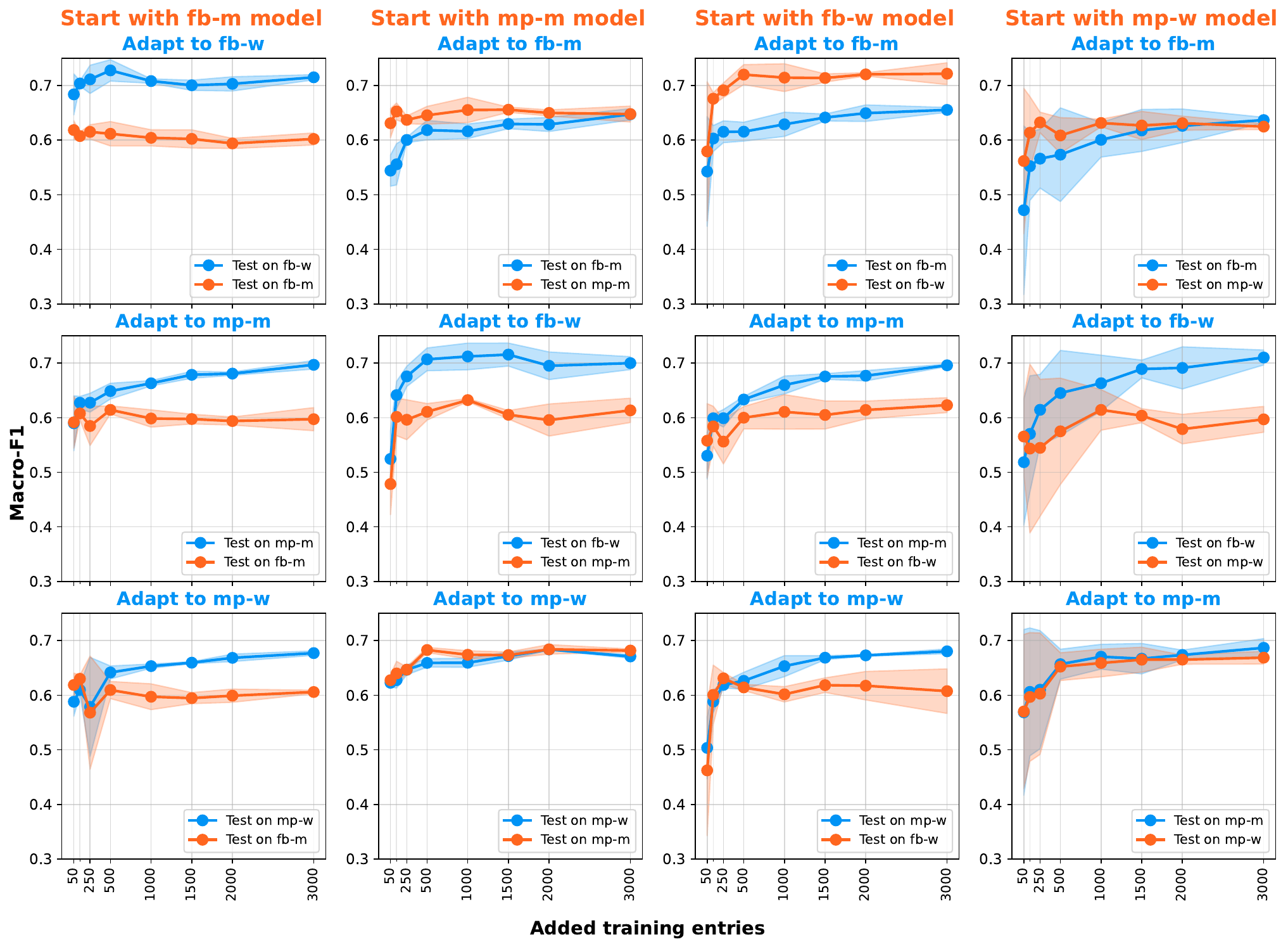}
    \caption{\small Learning curves for starting with a \texttt{dodo1} model trained on a single \texttt{dodo} pair and adding increments from the training set of a new \texttt{dodo} pair. We show mean and std-dev Macro-F1 (across 3 seeds) on the new adapt \texttt{dodo} and source start \texttt{dodo} at each increment.}
 \label{fig:double_learning_curves}
\end{figure*}

\begin{figure*}[!h]
    \centering
    \captionsetup{justification=raggedright}
    \includegraphics[width = 0.99\textwidth]{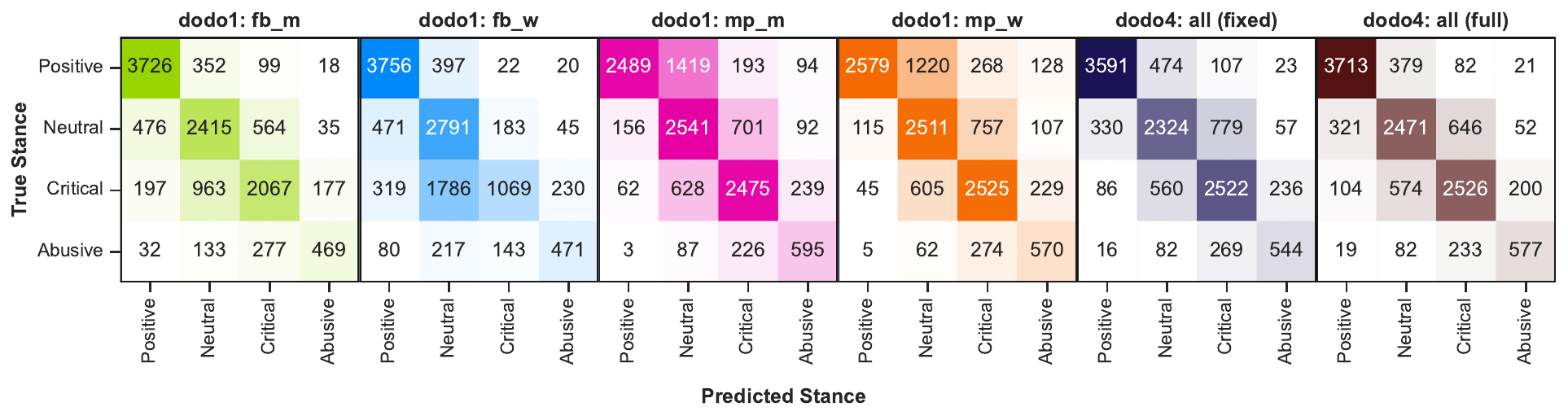}
    \caption{\small Confusion matrices for \texttt{dodo1} and \texttt{dodo4} models evaluated on the total test set (12,000 entries).}
 \label{fig:confusion}
\end{figure*}

\subsection{Only small amounts of data are needed to effectively adapt existing models to new domains and demographics.} 
\label{subsection:adaptation}
\vspace{-0.05em}
Here we \textit{start} with a fine-tuned specialist \texttt{dodo1} model (i.e., a model fine-tuned on a single \texttt{dodo}) and \textit{adapt} this model to a new \texttt{dodo}. We do continued fine-tuning  of each fine-tuned \texttt{dodo1} model on increments added from the adapt \texttt{dodo} train split.\footnote{The increments are [50, 125, 250, 500, 1000, 1500, 2000, 2500, 3000]. We train a separate model for each increment.} For the models trained using each budget increment, we calculate Macro-F1 on test sets of both the start and adaption \texttt{dodos} (see \cref{fig:double_learning_curves}) so that we record both performance gains in adapting to new \texttt{dodos} alongside performance losses (forgetting) in seen \texttt{dodos}. 

For almost all cases, the performance gain is notable after adding just 125 entries from the new \texttt{dodo} and increases with more entries. There is not a prominent performance gain after 500 entries except when adapting from fb-m to mp-m. This suggests that a small amount of data is efficient and cost-effective for testing how well existing models generalise. The importance of data composition over data quantity aligns with the fixed/full budget findings from \S\ref{section:smalldiverse}. On catastrophic forgetting, we generally do not find major performance drops. In some cases, adapting models to new data even helps classification in the source pair (e.g., mp-w to mp-m). Future work can explore where adaptation helps or hurts performance in source domains or demographics. %

\begin{figure}[!h]
    \centering
    \captionsetup{justification=justified}
    \includegraphics[width = 0.99\columnwidth]{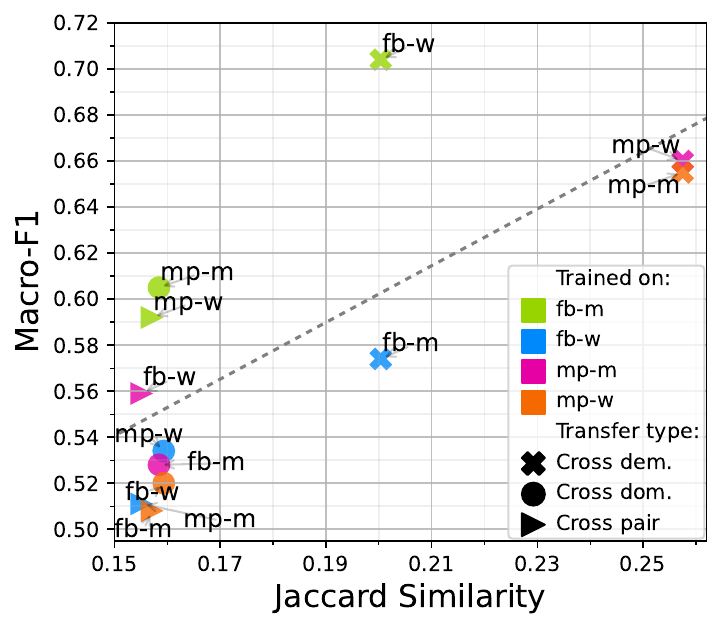}
    \caption{\small Jaccard similarity and mean 0-shot Macro-F1 for \texttt{dodo1} deBERT models with line of best fit. On graph annotations represent evaluation \texttt{dodo}. Shows positive correlation ($\rho=0.7$) and effectiveness of cross-demographic vs. cross-domain transfer.}
    \label{fig:scatter_similarity}
\end{figure}

\subsection{Dataset similarity is a signal of transferability.}
\label{section:sim_correlation}
Using the specialist \texttt{dodo1} models, we examine if dataset similarity signals transferability, i.e., the Macro-F1 score that a \texttt{dodo1} model can achieve on unseen \texttt{dodos}. We compute three classical text distance metrics with unigram bag-of-words approaches: Jaccard and S{\o}rensen-Dice similarity, and Kullback-Leibler divergence. In \cref{fig:scatter_similarity}, we plot Macro-F1 scores (of unseen single \texttt{dodos}) against Jaccard similarity for each pair of \texttt{dodos}. The correlation coefficient is 0.7, demonstrating a positive relationship between dataset similarity and unseen \texttt{dodo} performance.\footnote{Correlation coefficients are 0.7 for Dice Similarity and -0.66 for KL Divergence, confirming Jaccard robustness.} Greater similarity between demographic pairs versus domain pairs results in better cross-demographic transfer versus cross-domain transfer. Using these metrics could help estimate transfer potential before investing in an expensive labelling process.

\subsection{Error Analysis}

We find that errors made by \texttt{dodo1} models reflect the class imbalances outlined in \cref{sec:classdist}. We also see errors relating to inherent similarities across bordering classes, demonstrating the value of fine-grained labels. We present confusion matrices on the total test in \cref{fig:confusion}, and full error analysis in \cref{section:add_error_analysis}.
\section{Discussion}

We discuss the limitations of this work in \cref{sec:limitations}, addressing difficulties in disentangling the direction of sentiment in social media posts, limitations in the chosen label schema, and the consequences of the chosen evaluation approaches. Here, we present avenues for future work.

Expanding demographics and adding more complexity to the labelling schema would provide a broader basis for understanding generalisability in abuse classification. Other promising avenues include investigating whether active learning techniques \cite{vidgenTracking2022, kirkMoreDataBetter2022} aid more efficient cross-domain/demographic transfer, or whether architectures better suited for continual learning can assist in the addition of new groups without forgetting those previously trained-on \cite{huDrinkingFirehoseContinual2020, qianLifelongLearningHate2021a,liOvercomingCatastrophicForgetting2022}. We shuffled entries during training and used all four class labels but future work could assess whether performance is affected by order of training on different groups, and the impact of training on binary versus multi-class labels on transfer performance. Finally, our experiments only use fine-tuning on labelled data, but in-domain continued pre-training could be explored as a budget-efficient way to boost performance \cite{gururangan2020don, kirkSemEval2023}.

\section{Related Works}
\label{section:related_works}

\paragraph{Abuse Against MPs}
Academics and journalists account abuse against politicians, 
which may cause politicians to withdraw from their posts \cite{manningMPs2019, jamesAggressive2016}. %
Empirical work commonly studies Twitter  \cite{binnsThey2018, gorrellMP2020, wardTurds2020, agarwalHate2021}, including across national contexts such as European Parliament elections \cite{theocharisBad2016}, Canadian and US politicians \cite{rheaultPoliticians2019} and members of the UK parliament \cite{gorrellMP2020}.  
Other studies focus on gender differences in abuse \cite{rheaultPoliticians2019, eriksonThree2021} though some datasets only contain abuse against women \cite{stambolievaMethodology2017, delisleLargescale2019} which limits comparison across genders (unlike \textsc{DoDo}). %
Various techniques are employed to identify abusive tweets including rules-based or lexicon approaches and topic analysis \cite{gorrellTwits2018, gorrellMP2020, greenwoodOnline2019}; traditional machine learning classifiers \cite{stambolievaMethodology2017, rheaultPoliticians2019, agarwalHate2021} or pre-trained language models and off-the-shelf classifiers like Perspective API \cite{delisleLargescale2019}.

\paragraph{Abuse Against Footballers} Sport presents a good case for studying public figure abuse due to the influence of athletes \cite{carringtonIntroduction2012}, as well as the heightened symbolic focus on in-out groups and race-nation relations \cite{ brombergerFootball1995, backChanging2001, kingEuropean2003,burdseyRace2011, doidgeIf2015}. %
 Several studies track the change from racist chants at football stadiums, to the more pernicious and harder to control online abuse \cite{kingOffside2004, clelandRacism2013,  clelandFans2014, kilvingtonTackling2019}. %
Civil society organisations %
track social media abuse as far back as the 2012/2013 season, but are limited by a focus on manual case-by-case resolution and suffer from chronic underreporting  \cite{Bennett2017}. %
We build on our previous work in 
\citet{vidgenTracking2022},
which presents some of the same data as the male footballers portion in \textsc{DoDo} but also labels additional data using active learning.

\paragraph{Abuse Datasets and Detection}%
Developing robust abuse classifiers is challenging \citep{zhang2019hate}. Surveys on abuse detection cover various aspects such as algorithms \citep{schmidt-wiegand-2017-survey, mishra2019tackling}, model generalisability \citep{yin2021towards}, and data desiderata \citep{vidgenDirections2020}. %
Many %
studies curate data from mainstream platforms%
, focusing on abuse against different identities such as women \citep{fersini2018overview, PAMUNGKAS2020102360} and immigrants \citep{basile-etal-2019-semeval}. 
Recent approaches to developing abuse classifiers predominately fine-tune large language models on labelled datasets directly \citep{FORTUNA2021102524} (our approach) or in a multi-task setting \citep{waseem2018bridging, yuan2022detect}, as well as incorporate contextual information \citep{chiril2022emotionally}. Abuse detection datasets mostly focus on binary classification%
, and few cast the predictions as a multi-class problem. %
Some work addresses cross-domain classification in regards to generalisability \citep{glavasXHate9992020, yadav2023lahm, toraman_turkish, bourgeade_learn_to_hate, antypas_robust}, but many are either focused on combining existing datasets, or focus on domains as groups of content identified by keywords, as opposed to content sourced around members of a specific domain. 
The dataset we use in this paper rectifies some of these issues, containing fine-grained labels, and containing uniformly sourced and labelled content explicitly targeted at members of target groups.

\paragraph{Domain Adaptation}
Several NLP techniques have been explored for model generalisation in abuse detection, including feature-based domain alignment \citep{bashar2021progressive, ludwigImproving2022}, regularisation methods \citep{ludwigImproving2022}, and adaptive pre-training \citep{DBLP:conf/icaart/FaalYS21}. Systematic evaluation of model generalisability exists in some forms, focusing on dataset features \citep{FORTUNA2021102524}, multilinguality \citep{PAMUNGKAS2020102360, yadav2023lahm}, existing hate-speech datasets \citep{bourgeade_learn_to_hate}, and cross-domain generalisability where domains are keyword-based topics \citep{toraman_turkish}. To our knowledge there is no work that systemically explores the dynamics of transfer across both domain and demographic factors, using content specifically targeted at groups from different domains.

\section{Conclusion}

We fine-tuned language models using our \textsc{DoDo} dataset to classify abuse targeted at public figures for two domains (sports,  politics) and two demographics (women, men). We found that (i) even small amounts of diverse data provide significant benefits to generalisable performance and model adaptation; (ii) cross-demographic transfer (from women to men, or vice-versa) is more effective than cross-domain transfer (from footballers to MPs, or vice-versa) but models trained on data from one domain are less generalisable than models trained on cross-domain data; (iii) not all domains and demographics contribute equally to training generalisable models; and (iv) dataset similarity is a signal of transferability.

There are broader policy implications of our work. Policymakers, NGOs and others with an interest in independently monitoring harms face challenges in building models that are broad enough to capture a wide range of harms but specific enough to capture the distinctive nature of abuse (e.g., the difference between hate speech targeted at male and female MPs); while remaining within resource constraints typical of policy settings. Our work contributes by bringing fresh perspective on the feasibility of transferring models created to detect harm for one target to other targets. It thus provides insight into developing automated systems that are cost-effective, generalisable and performative across domains and demographics of interest. %

\section{Ethics and Harm Statement}
\label{sec:ethics}

We present our limitations section in \S\ref{sec:limitations}. In addition to these limitations, engaging with a subject such as online abuse raises ethical concerns. Here we set out the nature of those concerns, and how we managed them. Creation and annotation of a dataset focusing on abuse risks harming the annotators and researchers constructing the dataset, as repeated exposure to such material can be detrimental towards their mental health \cite{kirkHandling2022b}. Mitigating these risks is easier with a small trained team of annotators (like those we used for the MPs datasets) and harder with crowdworkers (like those we used for the footballers datasets). With the trained group of annotators, we maintained an open annotator forum where they could discuss such issues during the labelling process, and seek welfare support. For crowdworkers, we had very limited contact with them but include on our guidelines and task description extensive content warnings and links to publicly-available resources on vicarious trauma. 

We acknowledge that all experiments and data collection protocols are approved by the internal ethics review board at our institution.

\section{Limitations}
\label{sec:limitations}

\paragraph{Targets of Abuse} It is sometimes hard to disentangle the target of sentiment in tweets directed at public figures---some tweets praise public figures while simultaneously criticising another figure or even abusing identity groups (such as an praising an MP's anti-immigration policy while abusing immigrants). %
 Our label schema does not tag target-specific spans nor flag when it is a non-public figure account or abstract group is being abused. We also do not use further conversational context during annotation.  Furthermore, we are limited by gender distinctions in UK MPs statistics and football leagues---the dataset does not cover non-binary identities or other identity attributes. 

\paragraph{Types of Abuse} While our dataset is more diverse than most abuse datasets in including four class labels, it does not disaggregate abusive content into further subcategories such as identity attacks. Our preliminary keyword analysis suggested that identity attacks comprise a relatively small proportion of all abuse (especially for female footballers) but can nonetheless cause significant harm \cite{gelber_evidencing_2016}. Further investigation on abuse across demographic groups is needed to understand how women and men are targeted differently, and to assess distributional shifts of specific homophobic, racist, sexist or otherwise identity-based abuse.

\paragraph{Language and Platform Focus} Our dataset contains English language tweets associated with UK MPs and the top football leagues in England (though players come from a variety of nationalities). Prior studies suggest politicians face online abuse in other countries \citep{theocharisBad2016, ezeibeEthnic2017, rheaultPoliticians2019,  fuchsNormalizing2020, eriksonThree2021}; and that the English football social media audience is a global one \citep{kilvingtonTackling2019}. However, shifting national or cultural context will introduce further distributional and linguistic shifts. Furthermore, our data is only collected from Twitter though abuse towards public figures exists on a variety of social media platforms \cite{agarwalHate2021} such as YouTube \cite{espositoHow2021} or WhatsApp \cite{sahaShort2021}.

\paragraph{Evaluation Approach} 
Aggregate evaluation metrics may obscure per \texttt{dodo} and per class weaknesses \cite{rottger_hatecheck_2021}. The Macro-F1 score across the combined test set from all \texttt{dodos} does not equal the averaged Macro-F1 across each \texttt{dodo} test set (the former is 4.7pp higher on average). This is due to different class distributions across \texttt{dodos} skewing the total Macro-F1 calculation. The ranking of models was consistent across these two metrics. We have not investigated the relative dataset difficulty \cite{ethayarajh_understanding_2022} of individual \texttt{dodo} test sets, which may influence measures of generalisibility. 

\section*{Acknowledgements}
We would like to thank Dr Bertie Vidgen for prior direction and data annotation support, and Eirini Koutsouroupa for invaluable project management support. This work was supported by the Ecosystem Leadership Award under the EPSRC Grant EP\/X03870X\/1 \& The Alan Turing Institute. 

\nocite{*}
\section{Bibliographical References}\label{sec:reference}

\bibliographystyle{lrec-coling2024-natbib}
\bibliography{main}

\begin{thebibliography}{0}
\expandafter\ifx\csname natexlab\endcsname\relax\def\natexlab#1{#1}\fi

\end{thebibliography}


\begin{thebibliography}{96}
\expandafter\ifx\csname natexlab\endcsname\relax\def\natexlab#1{#1}\fi

\bibitem[{Agarwal et~al.(2021)Agarwal, Hawkins, Amaxopoulou, Dempsey, Sastry, and Wood}]{agarwalHate2021}
Pushkal Agarwal, Oliver Hawkins, Margarita Amaxopoulou, Noel Dempsey, Nishanth Sastry, and Edward Wood. 2021.
\newblock \href {https://doi.org/10.1145/3465336.3475113} {Hate {Speech} in {Political} {Discourse}: {A} {Case} {Study} of {UK} {MPs} on {Twitter}}.
\newblock \emph{Proceedings of the 32st ACM Conference on Hypertext and Social Media}, pages 5--16.
\newblock Publisher: ACM.

\bibitem[{Alkomah and Ma(2022)}]{alkomah_twitter_2022}
Fatimah Alkomah and Xiaogang Ma. 2022.
\newblock \href {https://doi.org/10.3390/info13060273} {A {Literature} {Review} of {Textual} {Hate} {Speech} {Detection} {Methods} and {Datasets}}.
\newblock \emph{Information}, 13(6):273.
\newblock Number: 6 Publisher: Multidisciplinary Digital Publishing Institute.

\bibitem[{Antypas and Camacho-Collados(2023)}]{antypas_robust}
Dimosthenis Antypas and Jose Camacho-Collados. 2023.
\newblock \href {https://doi.org/10.18653/v1/2023.woah-1.25} {Robust {Hate} {Speech} {Detection} in {Social} {Media}: {A} {Cross}-{Dataset} {Empirical} {Evaluation}}.
\newblock In \emph{The 7th {Workshop} on {Online} {Abuse} and {Harms} ({WOAH})}, pages 231--242, Toronto, Canada. Association for Computational Linguistics.

\bibitem[{Ayers et~al.(2018)Ayers, Caputi, Nebeker, and Dredze}]{ayers_twitter_2018}
John~W. Ayers, Theodore~L. Caputi, Camille Nebeker, and Mark Dredze. 2018.
\newblock \href {https://doi.org/10.1038/s41746-018-0036-2} {Don’t quote me: reverse identification of research participants in social media studies}.
\newblock \emph{npj Digital Medicine}, 1(1):1--2.
\newblock Number: 1 Publisher: Nature Publishing Group.

\bibitem[{Back et~al.(2001)Back, Crabbe, {John}, and Solomos}]{backChanging2001}
Les Back, Tim Crabbe, {John}, and John~Solomos Solomos. 2001.
\newblock \emph{The {Changing} {Face} of {Football}. {Racism}, {Identity} and {Multiculturc} in the {English} {Game}.}
\newblock Berg Publishers.

\bibitem[{Bardall(2013)}]{bardallGenderSpecific2013}
Gabrielle Bardall. 2013.
\newblock \href {https://doi.org/10.5334/sta.cs} {Gender-{Specific} {Election} {Violence}: {The} {Role} of {Information} and {Communication} {Technologies}}.
\newblock \emph{Stability: International Journal of Security \& Development}, 2(3):60.

\bibitem[{Bashar et~al.(2021)Bashar, Nayak, Luong, and Balasubramaniam}]{bashar2021progressive}
Md~Abul Bashar, Richi Nayak, Khanh Luong, and Thirunavukarasu Balasubramaniam. 2021.
\newblock Progressive domain adaptation for detecting hate speech on social media with small training set and its application to covid-19 concerned posts.
\newblock \emph{Social Network Analysis and Mining}, 11:1--18.

\bibitem[{Basile et~al.(2019)Basile, Bosco, Fersini, Nozza, Patti, Rangel~Pardo, Rosso, and Sanguinetti}]{basile-etal-2019-semeval}
Valerio Basile, Cristina Bosco, Elisabetta Fersini, Debora Nozza, Viviana Patti, Francisco~Manuel Rangel~Pardo, Paolo Rosso, and Manuela Sanguinetti. 2019.
\newblock \href {https://doi.org/10.18653/v1/S19-2007} {{S}em{E}val-2019 task 5: Multilingual detection of hate speech against immigrants and women in {T}witter}.
\newblock In \emph{Proceedings of the 13th International Workshop on Semantic Evaluation}, pages 54--63, Minneapolis, Minnesota, USA. Association for Computational Linguistics.

\bibitem[{Bender and Friedman(2018)}]{bender_friedman_2018}
Emily~M. Bender and Batya Friedman. 2018.
\newblock \href {https://doi.org/10.1162/tacl_a_00041} {Data statements for natural language processing: Toward mitigating system bias and enabling better science}.
\newblock \emph{Transactions of the Association for Computational Linguistics}, 6:587--604.

\bibitem[{Bennett and J{\"o}nsson(2017)}]{Bennett2017}
Hayley Bennett and Anna J{\"o}nsson. 2017.
\newblock Klick it out: tackling online discrimination in football.
\newblock In \emph{{Sport and Discrimination}}, page~12. Routledge.

\bibitem[{Bigi(2003)}]{modifed_KL}
Brigitte Bigi. 2003.
\newblock \href {https://doi.org/10.1007/3-540-36618-0\_22} {{Using Kullback-Leibler Distance for Text Categorization}}.
\newblock In \emph{{Advances in Information Retrieval}}, volume 2633, pages 305--319. {Springer Berlin Heidelberg}.

\bibitem[{Binns and Bateman(2018)}]{binnsThey2018}
Amy Binns and Martin Bateman. 2018.
\newblock \href {https://doi.org/10.1177/0956474818816860} {And they thought {Papers} were {Rude}}.
\newblock \emph{British Journalism Review}, 29(4):39--44.

\bibitem[{Bourgeade et~al.(2023)Bourgeade, Chiril, Benamara, and Moriceau}]{bourgeade_learn_to_hate}
Tom Bourgeade, Patricia Chiril, Farah Benamara, and Véronique Moriceau. 2023.
\newblock \href {https://doi.org/10.18653/v1/2023.eacl-main.254} {What {Did} {You} {Learn} {To} {Hate}? {A} {Topic}-{Oriented} {Analysis} of {Generalization} in {Hate} {Speech} {Detection}}.
\newblock In \emph{Proceedings of the 17th {Conference} of the {European} {Chapter} of the {Association} for {Computational} {Linguistics}}, pages 3495--3508, Dubrovnik, Croatia. Association for Computational Linguistics.

\bibitem[{Bright et~al.(2020)Bright, Hale, Ganesh, Bulovsky, Margetts, and Howard}]{bright2020campaigning}
Jonathan Bright, Scott Hale, Bharath Ganesh, Andrew Bulovsky, Helen Margetts, and Phil Howard. 2020.
\newblock \href {https://doi.org/10.1177/0093650219872394} {Does campaigning on social media make a difference? {Evidence} from candidate use of {Twitter} during the 2015 and 2017 {U.K.} elections}.
\newblock \emph{Communication Research}, 47(7):988--1009.

\bibitem[{Bromberger(1995)}]{brombergerFootball1995}
Christian Bromberger. 1995.
\newblock \href {https://doi.org/10.1177/095715589500601803} {Football as world-view and as ritual}.
\newblock \emph{French Cultural Studies}, 6(18):293--311.

\bibitem[{Brown(2009)}]{brownWWW2009}
Christopher Brown. 2009.
\newblock \href {https://doi.org/10.1080/10646170902869544} {{WWW}.{HATE}.{COM}: {White} {Supremacist} {Discourse} on the {Internet} and the {Construction} of {Whiteness} {Ideology}}.
\newblock \emph{Howard Journal of Communications}, 20(2):189--208.

\bibitem[{Burdsey(2011)}]{burdseyRace2011}
Daniel Burdsey. 2011.
\newblock \emph{Race, {Ethnicity} and {Football}}.
\newblock Routledge.

\bibitem[{Carrington(2012)}]{carringtonIntroduction2012}
Ben Carrington. 2012.
\newblock \href {https://doi.org/10.1080/01419870.2012.669488} {Introduction: sport matters}.
\newblock \emph{Ethnic and Racial Studies}, 35(6):961--970.

\bibitem[{Chiril et~al.(2022)Chiril, Pamungkas, Benamara, Moriceau, and Patti}]{chiril2022emotionally}
Patricia Chiril, Endang~Wahyu Pamungkas, Farah Benamara, V{\'e}ronique Moriceau, and Viviana Patti. 2022.
\newblock Emotionally informed hate speech detection: a multi-target perspective.
\newblock \emph{Cognitive Computation}, pages 1--31.

\bibitem[{Cleland(2013)}]{clelandRacism2013}
Jamie Cleland. 2013.
\newblock \href {https://doi.org/10.1177/0193723513499922} {Racism, {Football} {Fans}, and {Online} {Message} {Boards}}.
\newblock \emph{Journal of Sport and Social Issues}, 38(5):415--431.
\newblock Publisher: SAGE PublicationsSage CA: Los Angeles, CA.

\bibitem[{Cleland and Cashmore(2014)}]{clelandFans2014}
Jamie Cleland and Ellis Cashmore. 2014.
\newblock \href {https://doi.org/10.1080/1369183X.2013.777524} {Fans, {Racism} and {British} {Football} in the {Twenty}-{First} {Century}: {The} {Existence} of a `{Colour}-{Blind}' {Ideology}}.
\newblock \emph{Journal of Ethnic and Migration Studies}, 40(4):638--654.

\bibitem[{Colchester(2022)}]{colchester_boris_2022}
Max Colchester. 2022.
\newblock \href {https://www.wsj.com/articles/boris-johnson-apologizes-for-attending-party-at-downing-street-during-lockdown-11641994454} {Boris {Johnson} {Apologizes} for {Party} at {Downing} {Street} {During} {U}.{K}. {Lockdown}}.
\newblock \emph{Wall Street Journal}.

\bibitem[{Coleman(1999)}]{colemanCan1999}
Stephen Coleman. 1999.
\newblock \href {https://doi.org/10.1111/1467-923X.00200} {Can the {New} {Media} {Invigorate} {Democracy}?}
\newblock \emph{The Political Quarterly}, 70(1):16--22.

\bibitem[{Coleman(2005)}]{colemanNew2005}
Stephen Coleman. 2005.
\newblock \href {https://doi.org/10.1177/1461444805050745} {New mediation and direct representation: reconceptualizing representation in the digital age}.
\newblock \emph{New Media \& Society}, 7(2):177--198.

\bibitem[{Coleman and Spiller(2003)}]{colemanExploring2003}
Stephen Coleman and Josephine Spiller. 2003.
\newblock \href {https://doi.org/10.1080/1357233042000246837} {Exploring new media effects on representative democracy}.
\newblock \emph{The Journal of Legislative Studies}, 9(3):1--16.

\bibitem[{Davidson et~al.(2017)Davidson, Warmsley, Macy, and Weber}]{davidsonAutomated2017}
Thomas Davidson, Dana Warmsley, Michael Macy, and Ingmar Weber. 2017.
\newblock \href {http://arxiv.org/abs/1703.04009} {Automated {Hate} {Speech} {Detection} and the {Problem} of {Offensive} {Language}}.
\newblock \emph{Proceedings of the 11th International Conference on Web and Social Media, ICWSM 2017}, pages 512--515.
\newblock Publisher: AAAI Press.

\bibitem[{Delisle et~al.(2019)Delisle, Kalaitzis, Majewski, de~Berker, Marin, and Cornebise}]{delisleLargescale2019}
Laure Delisle, Alfredo Kalaitzis, Krzysztof Majewski, Archy de~Berker, Milena Marin, and Julien Cornebise. 2019.
\newblock \href {https://doi.org/10.48550/ARXIV.1902.03093} {A large-scale crowdsourced analysis of abuse against women journalists and politicians on twitter}.

\bibitem[{Dewey(1927)}]{deweyPublic1927}
John Dewey. 1927.
\newblock \emph{The public and its problem}.
\newblock Henry Holt.

\bibitem[{Doidge(2015)}]{doidgeIf2015}
Mark Doidge. 2015.
\newblock \href {https://doi.org/10.1177/1012690213480354} {`{If} you jump up and down, {Balotelli} dies': {Racism} and player abuse in {Italian} football}.
\newblock \emph{International Review for the Sociology of Sport}, 50(3):249--264.
\newblock Publisher: SAGE PublicationsSage UK: London, England.

\bibitem[{ElSherief et~al.(2018)ElSherief, Kulkarni, Nguyen, Wang, and Belding}]{elsherief2018hate}
Mai ElSherief, Vivek Kulkarni, Dana Nguyen, William~Yang Wang, and Elizabeth Belding. 2018.
\newblock Hate lingo: A target-based linguistic analysis of hate speech in social media.
\newblock In \emph{Proceedings of the International AAAI Conference on Web and Social Media}, pages 42--51.

\bibitem[{Erikson et~al.(2021)Erikson, H{\aa}kansson, and Josefsson}]{eriksonThree2021}
Josefina Erikson, Sandra H{\aa}kansson, and Cecilia Josefsson. 2021.
\newblock \href {https://doi.org/10.1017/S1537592721002048} {Three {Dimensions} of {Gendered} {Online} {Abuse}: {Analyzing} {Swedish} {MPs}' {Experiences} of {Social} {Media}}.
\newblock \emph{Perspectives on Politics}, pages 1--17.
\newblock Publisher: Cambridge University Press.

\bibitem[{Esposito and Zollo(2021)}]{espositoHow2021}
Eleonora Esposito and Sole~Alba Zollo. 2021.
\newblock \href {https://doi.org/10.1075/jlac.00053.esp} {``{How} dare you call her a pig, {I} know several pigs who would be upset if they knew''*}.
\newblock \emph{Journal of Language Aggression and Conflict}, 9(1):47--75.

\bibitem[{Ethayarajh et~al.(2022)Ethayarajh, Choi, and Swayamdipta}]{ethayarajh_understanding_2022}
Kawin Ethayarajh, Yejin Choi, and Swabha Swayamdipta. 2022.
\newblock \href {https://proceedings.mlr.press/v162/ethayarajh22a.html} {Understanding {Dataset} {Difficulty} with \${\textbackslash}mathcal\{{V}\}\$-{Usable} {Information}}.
\newblock In \emph{Proceedings of the 39th {International} {Conference} on {Machine} {Learning}}, pages 5988--6008. PMLR.
\newblock ISSN: 2640-3498.

\bibitem[{Ezeibe and Ikeanyibe(2017)}]{ezeibeEthnic2017}
Christian~Chukwuebuka Ezeibe and Okey~Marcellus Ikeanyibe. 2017.
\newblock \href {https://doi.org/10.2979/africatoday.63.4.04} {Ethnic {Politics}, {Hate} {Speech}, and {Access} to {Political} {Power} in {Nigeria}}.
\newblock \emph{Africa Today}, 63(4):65.

\bibitem[{Faal et~al.(2021)Faal, Yu, and Schmitt}]{DBLP:conf/icaart/FaalYS21}
Farshid Faal, Jia~Yuan Yu, and Ketra~A. Schmitt. 2021.
\newblock \href {https://doi.org/10.5220/0010266109320940} {Domain adaptation multi-task deep neural network for mitigating unintended bias in toxic language detection}.
\newblock In \emph{Proceedings of the 13th International Conference on Agents and Artificial Intelligence, {ICAART} 2021, Volume 2, Online Streaming, February 4-6, 2021}, pages 932--940. {SCITEPRESS}.

\bibitem[{Farrington et~al.(2014)Farrington, Hall, Kilvington, Price, and Saeed}]{farringtonSport2014}
N.~Farrington, L.~Hall, D.~Kilvington, J.~Price, and A.~Saeed. 2014.
\newblock \emph{Sport, racism and social media}.
\newblock Routledge.

\bibitem[{Fersini et~al.(2018)Fersini, Nozza, Rosso et~al.}]{fersini2018overview}
Elisabetta Fersini, Debora Nozza, Paolo Rosso, et~al. 2018.
\newblock Overview of the evalita 2018 task on automatic misogyny identification (ami).
\newblock In \emph{EVALITA Evaluation of NLP and Speech Tools for Italian Proceedings of the Final Workshop 12-13 December 2018, Naples}. Accademia University Press.

\bibitem[{Fortuna et~al.(2021)Fortuna, Soler-Company, and Wanner}]{FORTUNA2021102524}
Paula Fortuna, Juan Soler-Company, and Leo Wanner. 2021.
\newblock \href {https://doi.org/https://doi.org/10.1016/j.ipm.2021.102524} {How well do hate speech, toxicity, abusive and offensive language classification models generalize across datasets?}
\newblock \emph{Information Processing \& Management}, 58(3):102524.

\bibitem[{Frosdick and Marsh(2013)}]{football_hooliganism}
Steve Frosdick and Peter Marsh. 2013.
\newblock \emph{Football {Hooliganism}}.
\newblock Routledge.

\bibitem[{Fuchs and Sch{\"a}fer(2020)}]{fuchsNormalizing2020}
Tamara Fuchs and Fabian Sch{\"a}fer. 2020.
\newblock \href {https://doi.org/10.1080/09555803.2019.1687564} {Normalizing misogyny: hate speech and verbal abuse of female politicians on {Japanese} {Twitter}}.
\newblock \emph{Japan Forum}, pages 1--27.

\bibitem[{Gelber and McNamara(2016)}]{gelber_evidencing_2016}
Katharine Gelber and Luke McNamara. 2016.
\newblock \href {https://doi.org/10.1080/13504630.2015.1128810} {Evidencing the harms of hate speech}.
\newblock \emph{Social Identities}, 22(3):324--341.
\newblock Publisher: Routledge \_eprint: https://doi.org/10.1080/13504630.2015.1128810.

\bibitem[{Glava{\v s} et~al.(2020)Glava{\v s}, Karan, and Vuli{\'c}}]{glavasXHate9992020}
Goran Glava{\v s}, Mladen Karan, and Ivan Vuli{\'c}. 2020.
\newblock \href {https://doi.org/10.18653/v1/2020.coling-main.559} {{XHate}-999: {Analyzing} and {Detecting} {Abusive} {Language} {Across} {Domains} and {Languages}}.
\newblock In \emph{Proceedings of the 28th {International} {Conference} on {Computational} {Linguistics}}, pages 6350--6365, Barcelona, Spain (Online). International Committee on Computational Linguistics.

\bibitem[{Gorrell et~al.(2020)Gorrell, Farrell, and Bontcheva}]{gorrellMP2020}
Genevieve Gorrell, Tracie Farrell, and Kalina Bontcheva. 2020.
\newblock \href {https://doi.org/10.48550/ARXIV.2006.08363} {Mp twitter abuse in the age of covid-19: White paper}.

\bibitem[{Gorrell et~al.(2018)Gorrell, Greenwood, Roberts, Maynard, and Bontcheva}]{gorrellTwits2018}
Genevieve Gorrell, Mark Greenwood, Ian Roberts, Diana Maynard, and Kalina Bontcheva. 2018.
\newblock \href {https://doi.org/10.1609/icwsm.v12i1.15070} {Twits, {Twats} and {Twaddle}: {Trends} in {Online} {Abuse} towards {UK} {Politicians} and twaddle: Trends in online abuse towards {UK} politicians}.
\newblock \emph{Proceedings of the International {AAAI} Conference on Web and Social Media}, 12(1).

\bibitem[{Greenwood et~al.(2019)Greenwood, Bakir, Gorrell, Song, Roberts, and Bontcheva}]{greenwoodOnline2019}
Mark~A. Greenwood, Mehmet~E. Bakir, Genevieve Gorrell, Xingyi Song, Ian Roberts, and Kalina Bontcheva. 2019.
\newblock \href {https://doi.org/10.48550/ARXIV.1904.11230} {Online abuse of uk mps from 2015 to 2019: Working paper}.

\bibitem[{Gururangan et~al.(2020)Gururangan, Marasovi{\'c}, Swayamdipta, Lo, Beltagy, Downey, and Smith}]{gururangan2020don}
Suchin Gururangan, Ana Marasovi{\'c}, Swabha Swayamdipta, Kyle Lo, Iz~Beltagy, Doug Downey, and Noah~A Smith. 2020.
\newblock Don't stop pretraining: Adapt language models to domains and tasks.
\newblock In \emph{Proceedings of the 58th Annual Meeting of the Association for Computational Linguistics}, pages 8342--8360.

\bibitem[{Gwet(2014)}]{gwet_handbook_2014}
Kilem~L. Gwet. 2014.
\newblock \emph{Handbook of inter-rater reliability: {The} definitive guide to measuring the extent of agreement among raters}.
\newblock Advanced Analytics, LLC.

\bibitem[{He et~al.(2021)He, Gao, and Chen}]{he2021debertav3}
Pengcheng He, Jianfeng Gao, and Weizhu Chen. 2021.
\newblock Debertav3: Improving deberta using electra-style pre-training with gradient-disentangled embedding sharing.
\newblock \emph{arXiv preprint arXiv:2111.09543}.

\bibitem[{Holden and Phillips(2021)}]{football_euro_abuse}
Michael Holden and Mitch Phillips. 2021.
\newblock \href {https://www.reuters.com/world/uk/uk-pm-johnson-condemns- racist-abuse-england-soccer-team-2021-07-12/} {England's {Black} players face racial abuse after {Euro} 2020 defeat}.
\newblock \emph{Reuters}.

\bibitem[{Hu et~al.(2020)Hu, Sener, Sha, and Koltun}]{huDrinkingFirehoseContinual2020}
Hexiang Hu, Ozan Sener, Fei Sha, and Vladlen Koltun. 2020.
\newblock \href {http://arxiv.org/abs/2007.09335} {Drinking from a firehose: Continual learning with web-scale natural language}.
\newblock Version: 2.

\bibitem[{Ingle(2021)}]{ingleSports2021}
Sean Ingle. 2021.
\newblock \href {https://www.theguardian.com/sport/2021/apr/29/major-sports- bodies-84-hour-social-media-boycott-over-online-abuse- facebook-twitter} {Sports bodies to boycott social media for bank holiday weekend over abuse}.
\newblock \emph{The Guardian}.

\bibitem[{James et~al.(2016)James, Farnham, Sukhwal, Jones, Carlisle, and Henley}]{jamesAggressive2016}
David~V. James, Frank~R. Farnham, Seema Sukhwal, Katherine Jones, Josephine Carlisle, and Sara Henley. 2016.
\newblock \href {https://doi.org/10.1080/14789949.2015.1124908} {Aggressive/intrusive behaviours, harassment and stalking of members of the {United} {Kingdom} parliament: a prevalence study and cross-national comparison}.
\newblock \emph{The Journal of Forensic Psychiatry \& Psychology}, 27(2):177--197.

\bibitem[{Joinson et~al.(2009)Joinson, McKenna, Postmes, and Reips}]{joinsonOxford2012}
Adam Joinson, Katelyn Y.~A. McKenna, Tom Postmes, and Ulf-Dietrich Reips. 2009.
\newblock \href {https://doi.org/10.1093/oxfordhb/9780199561803.001.0001} {\emph{{Oxford Handbook of Internet Psychology}}}.
\newblock Oxford University Press.

\bibitem[{Kilvington and Price(2019)}]{kilvingtonTackling2019}
Daniel Kilvington and John Price. 2019.
\newblock \href {https://doi.org/10.1177/2167479517745300} {Tackling {Social} {Media} {Abuse}? {Critically} {Assessing} {English} {Football}'s {Response} to {Online} {Racism}}.
\newblock \emph{Communication \& Sport}, 7(1):64--79.
\newblock Publisher: SAGE PublicationsSage CA: Los Angeles, CA.

\bibitem[{King(2003)}]{kingEuropean2003}
Anthony King. 2003.
\newblock \emph{The {European} {Ritual}: {Football} in the {New} {Europe}}.
\newblock Ashgate Publishing Ltd.

\bibitem[{King(2004)}]{kingOffside2004}
Colin King. 2004.
\newblock \emph{Offside racism: {Playing} the white man}.
\newblock Routledge.

\bibitem[{Kirk et~al.(2022{\natexlab{a}})Kirk, Birhane, Vidgen, and Derczynski}]{kirkHandling2022b}
Hannah Kirk, Abeba Birhane, Bertie Vidgen, and Leon Derczynski. 2022{\natexlab{a}}.
\newblock \href {https://aclanthology.org/2022.findings-emnlp.35} {Handling and {Presenting} {Harmful} {Text} in {NLP} {Research}}.
\newblock In \emph{Findings of the {Association} for {Computational} {Linguistics}: {EMNLP} 2022}, pages 497--510, Abu Dhabi, United Arab Emirates. Association for Computational Linguistics.

\bibitem[{Kirk et~al.(2022{\natexlab{b}})Kirk, Vidgen, R{\"o}ttger, Thrush, and Hale}]{kirk2022hatemoji}
Hannah Kirk, Bertie Vidgen, Paul R{\"o}ttger, Tristan Thrush, and Scott Hale. 2022{\natexlab{b}}.
\newblock Hatemoji: A test suite and adversarially-generated dataset for benchmarking and detecting emoji-based hate.
\newblock In \emph{Proceedings of the 2022 Conference of the North American Chapter of the Association for Computational Linguistics: Human Language Technologies}, pages 1352--1368.

\bibitem[{Kirk et~al.(2022{\natexlab{c}})Kirk, Vidgen, and Hale}]{kirkMoreDataBetter2022}
Hannah~Rose Kirk, Bertie Vidgen, and Scott~A. Hale. 2022{\natexlab{c}}.
\newblock \href {https://arxiv.org/abs/2209.10193} {Is {More} {Data} {Better}? {Using} {Transformers}-{Based} {Active} {Learning} for {Efficient} and {Effective} {Detection} of {Abusive} {Language}}.
\newblock In \emph{Proceedings of the 3rd workshop on {Threat}, {Aggression} and {Cyberbullying} ({COLING} 2022)}. Association for Computational Linguistics.

\bibitem[{Kirk et~al.(2023)Kirk, Yin, Vidgen, and R{\"o}ttger}]{kirkSemEval2023}
Hannah~Rose Kirk, Wenjie Yin, Bertie Vidgen, and Paul R{\"o}ttger. 2023.
\newblock \href {https://doi.org/10.48550/arXiv.2303.04222} {{SemEval}-2023 {Task} 10: {Explainable} {Detection} of {Online} {Sexism}}.
\newblock In \emph{Proceedings of the 17th International Workshop on Semantic Evaluation}, Toronto, Canada. {Association for Computational Linguistics}.

\bibitem[{Li et~al.(2022)Li, Chen, Cho, Hao, Liu, Xing, Guo, and Liu}]{liOvercomingCatastrophicForgetting2022}
Dingcheng Li, Zheng Chen, Eunah Cho, Jie Hao, Xiaohu Liu, Fan Xing, Chenlei Guo, and Yang Liu. 2022.
\newblock \href {https://aclanthology.org/2022.naacl-main.398} {Overcoming catastrophic forgetting during domain adaptation of seq2seq language generation}.
\newblock In \emph{Proceedings of the 2022 Conference of the North American Chapter of the Association for Computational Linguistics: Human Language Technologies}, pages 5441--5454. Association for Computational Linguistics.

\bibitem[{Ludwig et~al.(2022)Ludwig, Dolos, Zesch, and Hobley}]{ludwigImproving2022}
Florian Ludwig, Klara Dolos, Torsten Zesch, and Eleanor Hobley. 2022.
\newblock \href {https://doi.org/10.18653/v1/2022.woah-1.4} {Improving {Generalization} of {Hate} {Speech} {Detection} {Systems} to {Novel} {Target} {Groups} via {Domain} {Adaptation}}.
\newblock In \emph{Proceedings of the {Sixth} {Workshop} on {Online} {Abuse} and {Harms} ({WOAH})}, pages 29--39, Seattle, Washington (Hybrid). Association for Computational Linguistics.

\bibitem[{Lynch et~al.(2022)Lynch, Sherlock, and Bradshaw}]{mp_abuse}
Paul Lynch, Pete Sherlock, and Paul Bradshaw. 2022.
\newblock \href {https://www.bbc.com/news/uk-63330885} {Scale of abuse of politicians on {Twitter} revealed}.
\newblock \emph{BBC News}.

\bibitem[{Manning and Kemp(2019)}]{manningMPs2019}
Lucy Manning and Phillip Kemp. 2019.
\newblock \href {https://www.bbc.com/news/uk-politics-49247808} {{MPs} describe threats, abuse and safety fears}.
\newblock \emph{BBC News}.

\bibitem[{Meloy et~al.(2008)Meloy, Sheridan, and Hoffmann}]{meloyStalking2008}
J.~Reid Meloy, Lorraine Sheridan, and Jens Hoffmann, editors. 2008.
\newblock \href {https://doi.org/10.1093/med:psych/9780195326383.001.0001} {\emph{Stalking, {Threatening}, and {Attacking} {Public} {Figures}}}.
\newblock Oxford University Press.

\bibitem[{Mishra et~al.(2019)Mishra, Yannakoudakis, and Shutova}]{mishra2019tackling}
Pushkar Mishra, Helen Yannakoudakis, and Ekaterina Shutova. 2019.
\newblock Tackling online abuse: A survey of automated abuse detection methods.
\newblock \emph{arXiv preprint arXiv:1908.06024}.

\bibitem[{Mullen et~al.(2009)Mullen, James, Meloy, Path{\'e}, Farnham, Preston, Darnley, and Berman}]{mullenFixated2009}
Paul~E. Mullen, David~V. James, J.~Reid Meloy, Michele~T. Path{\'e}, Frank~R. Farnham, Lulu Preston, Brian Darnley, and Jeremy Berman. 2009.
\newblock \href {https://doi.org/10.1080/14789940802197074} {The fixated and the pursuit of public figures}.
\newblock \emph{Journal of Forensic Psychiatry \& Psychology}, 20(1):33--47.
\newblock Publisher: Routledge.

\bibitem[{Norris(1999)}]{norrisCritical1999}
Pippa Norris. 1999.
\newblock \href {https://doi.org/10.1093/0198295685.001.0001} {\emph{Critical {Citizens}}}.
\newblock Oxford University Press.

\bibitem[{Pamungkas et~al.(2020)Pamungkas, Basile, and Patti}]{PAMUNGKAS2020102360}
Endang~Wahyu Pamungkas, Valerio Basile, and Viviana Patti. 2020.
\newblock \href {https://doi.org/https://doi.org/10.1016/j.ipm.2020.102360} {Misogyny detection in twitter: a multilingual and cross-domain study}.
\newblock \emph{Information Processing \& Management}, 57(6):102360.

\bibitem[{Papacharissi(2004)}]{papacharissi2004democracy}
Zizi Papacharissi. 2004.
\newblock Democracy online: Civility, politeness, and the democratic potential of online political discussion groups.
\newblock \emph{New media \& society}, 6(2):259--283.

\bibitem[{Parliament(2010)}]{EqualityAct2010}
{UK} Parliament. 2010.
\newblock \href {https://www.legislation.gov.uk/ukpga/2010/15/contents} {{E}quality {A}ct 2010}.

\bibitem[{Qian et~al.(2021)Qian, Wang, ElSherief, and Yan}]{qianLifelongLearningHate2021a}
Jing Qian, Hong Wang, Mai ElSherief, and Xifeng Yan. 2021.
\newblock \href {https://doi.org/10.48550/arXiv.2106.02821} {Lifelong {Learning} of {Hate} {Speech} {Classification} on {Social} {Media}}.
\newblock ArXiv:2106.02821 [cs].

\bibitem[{Rheault et~al.(2019)Rheault, Rayment, and Musulan}]{rheaultPoliticians2019}
Ludovic Rheault, Erica Rayment, and Andreea Musulan. 2019.
\newblock \href {https://doi.org/10.1177/2053168018816228} {Politicians in the line of fire: {Incivility} and the treatment of women on social media}.
\newblock \emph{Research \& Politics}, 6(1).

\bibitem[{Rowe(2015)}]{roweCivility2015}
Ian Rowe. 2015.
\newblock \href {https://doi.org/10.1080/1369118X.2014.940365} {Civility 2.0: a comparative analysis of incivility in online political discussion}.
\newblock \emph{Information, Communication \& Society}, 18(2):121--138.

\bibitem[{Röttger et~al.(2021)Röttger, Vidgen, Nguyen, Waseem, Margetts, and Pierrehumbert}]{rottger_hatecheck_2021}
Paul Röttger, Bertie Vidgen, Dong Nguyen, Zeerak Waseem, Helen Margetts, and Janet Pierrehumbert. 2021.
\newblock \href {https://doi.org/10.18653/v1/2021.acl-long.4} {{HateCheck}: {Functional} {Tests} for {Hate} {Speech} {Detection} {Models}}.
\newblock In \emph{Proceedings of the 59th {Annual} {Meeting} of the {Association} for {Computational} {Linguistics} and the 11th {International} {Joint} {Conference} on {Natural} {Language} {Processing} ({Volume} 1: {Long} {Papers})}, pages 41--58, Online. Association for Computational Linguistics.

\bibitem[{Saha et~al.(2021)Saha, Mathew, Garimella, and Mukherjee}]{sahaShort2021}
Punyajoy Saha, Binny Mathew, Kiran Garimella, and Animesh Mukherjee. 2021.
\newblock \href {https://doi.org/10.1145/3442381.3450137} {``{Short} is the {Road} that {Leads} from {Fear} to {Hate}'': {Fear} {Speech} in {Indian} {WhatsApp} {Groups}}.
\newblock In \emph{Proceedings of the {Web} {Conference} 2021}, pages 1110--1121. ACM.

\bibitem[{Sanh et~al.(2019)Sanh, Debut, Chaumond, and Wolf}]{sanh2019distilbert}
Victor Sanh, Lysandre Debut, Julien Chaumond, and Thomas Wolf. 2019.
\newblock Distilbert, a distilled version of bert: smaller, faster, cheaper and lighter.
\newblock \emph{arXiv preprint arXiv:1910.01108}.

\bibitem[{Schmidt and Wiegand(2017)}]{schmidt-wiegand-2017-survey}
Anna Schmidt and Michael Wiegand. 2017.
\newblock \href {https://doi.org/10.18653/v1/W17-1101} {A survey on hate speech detection using natural language processing}.
\newblock In \emph{Proceedings of the Fifth International Workshop on Natural Language Processing for Social Media}, pages 1--10, Valencia, Spain. Association for Computational Linguistics.

\bibitem[{Shulman(2009)}]{shulmanCase2009}
Stuart~W. Shulman. 2009.
\newblock \href {https://doi.org/10.2202/1944-2866.1010} {The {Case} {Against} {Mass} {E}-mails: {Perverse} {Incentives} and {Low} {Quality} {Public} {Participation} in {U}.{S}. {Federal} {Rulemaking}}.
\newblock \emph{Policy \& Internet}, 1(1):22--52.

\bibitem[{Stambolieva(2017)}]{stambolievaMethodology2017}
Ekaterina Stambolieva. 2017.
\newblock Methodology : {Detecting} {Online} {Abuse} against {Women} {MPs} on {Twitter}.
\newblock Technical Report Amnesty International, Amnesty International.

\bibitem[{Suler(2004)}]{sulerOnline2004}
John Suler. 2004.
\newblock \href {https://doi.org/10.1089/1094931041291295} {The {Online} {Disinhibition} {Effect}}.
\newblock \emph{CyberPsychology \& Behavior}, 7(3):321--326.

\bibitem[{Talat et~al.(2018)Talat, Thorne, and Bingel}]{waseem2018bridging}
Zeerak Talat, James Thorne, and Joachim Bingel. 2018.
\newblock Bridging the gaps: Multi task learning for domain transfer of hate speech detection.
\newblock \emph{Online harassment}, pages 29--55.

\bibitem[{Theocharis et~al.(2016)Theocharis, Barber{\'a}, Fazekas, Popa, and Parnet}]{theocharisBad2016}
Yannis Theocharis, Pablo Barber{\'a}, Zolt{\'a}n Fazekas, Sebastian~Adrian Popa, and Olivier Parnet. 2016.
\newblock \href {https://doi.org/10.1111/jcom.12259} {A {Bad} {Workman} {Blames} {His} {Tweets}: {The} {Consequences} of {Citizens}' {Uncivil} {Twitter} {Use} {When} {Interacting} {With} {Party} {Candidates}}.
\newblock \emph{Journal of Communication}, 66(6):1007--1031.

\bibitem[{Toraman et~al.(2022)Toraman, Şahinuç, and Yilmaz}]{toraman_turkish}
Cagri Toraman, Furkan Şahinuç, and Eyup Yilmaz. 2022.
\newblock \href {https://aclanthology.org/2022.lrec-1.238} {Large-{Scale} {Hate} {Speech} {Detection} with {Cross}-{Domain} {Transfer}}.
\newblock In \emph{Proceedings of the {Thirteenth} {Language} {Resources} and {Evaluation} {Conference}}, pages 2215--2225, Marseille, France. European Language Resources Association.

\bibitem[{Vidgen et~al.(2022)Vidgen, Chung, Johansson, Kirk, Williams, Hale, Margetts, R{\"o}ttger, and Sprejer}]{vidgenTracking2022}
Bertie Vidgen, Yi-Ling Chung, Pica Johansson, Hannah~Rose Kirk, Angus Williams, Scott~A. Hale, Helen~Zerlina Margetts, Paul R{\"o}ttger, and Laila Sprejer. 2022.
\newblock \href {https://doi.org/10.2139/ssrn.4403913} {Tracking {Abuse} on {Twitter} {Against} {Football} {Players} in the 2021 -- 22 {Premier} {League} {Season}}.

\bibitem[{Vidgen and Derczynski(2020)}]{vidgenDirections2020}
Bertie Vidgen and Leon Derczynski. 2020.
\newblock \href {https://doi.org/10.1371/journal.pone.0243300} {Directions in abusive language training data, a systematic review: {Garbage} in, garbage out}.
\newblock \emph{PLOS ONE}, 15(12):e0243300.
\newblock Publisher: Public Library of Science.

\bibitem[{Vidgen et~al.(2019)Vidgen, Harris, Nguyen, Tromble, Hale, and Margetts}]{vidgen2019challenges}
Bertie Vidgen, Alex Harris, Dong Nguyen, Rebekah Tromble, Scott Hale, and Helen Margetts. 2019.
\newblock Challenges and frontiers in abusive content detection.
\newblock In \emph{Proceedings of the Third Workshop on Abusive Language Online}, pages 80--93.

\bibitem[{Vidgen et~al.(2021{\natexlab{a}})Vidgen, Nguyen, Margetts, Rossini, and Tromble}]{vidgenIntroducing2021}
Bertie Vidgen, Dong Nguyen, Helen Margetts, Patricia Rossini, and Rebekah Tromble. 2021{\natexlab{a}}.
\newblock \href {https://doi.org/10.18653/v1/2021.naacl-main.182} {Introducing {CAD}: the contextual abuse dataset}.
\newblock In \emph{Proceedings of the 2021 Conference of the North American Chapter of the Association for Computational Linguistics: Human Language Technologies}, pages 2289--2303, Online. Association for Computational Linguistics.

\bibitem[{Vidgen et~al.(2021{\natexlab{b}})Vidgen, Thrush, Waseem, and Kiela}]{vidgen2021learning}
Bertie Vidgen, Tristan Thrush, Zeerak Waseem, and Douwe Kiela. 2021{\natexlab{b}}.
\newblock Learning from the worst: Dynamically generated datasets to improve online hate detection.
\newblock In \emph{Proceedings of the 59th Annual Meeting of the Association for Computational Linguistics and the 11th International Joint Conference on Natural Language Processing (Volume 1: Long Papers)}, pages 1667--1682.

\bibitem[{Ward and McLoughlin(2020)}]{wardTurds2020}
Stephen Ward and Liam McLoughlin. 2020.
\newblock \href {https://doi.org/10.1080/13572334.2020.1730502} {Turds, traitors and tossers: the abuse of {UK} {MPs} via {Twitter}}.
\newblock \emph{The Journal of Legislative Studies}, 26(1):47--73.

\bibitem[{Williamson(2009)}]{williamsonEffect2009}
Andy Williamson. 2009.
\newblock \href {https://doi.org/10.1093/pa/gsp009} {The {Effect} of {Digital} {Media} on {MPs}' {Communication} with {Constituents}}.
\newblock \emph{Parliamentary Affairs}, 62(3):514--527.

\bibitem[{Wolf et~al.(2020)Wolf, Debut, Sanh, Chaumond, Delangue, Moi, Cistac, Rault, Louf, Funtowicz, Davison, Shleifer, von Platen, Ma, Jernite, Plu, Xu, Le~Scao, Gugger, Drame, Lhoest, and Rush}]{huggingface_transformers}
Thomas Wolf, Lysandre Debut, Victor Sanh, Julien Chaumond, Clement Delangue, Anthony Moi, Pierric Cistac, Tim Rault, Remi Louf, Morgan Funtowicz, Joe Davison, Sam Shleifer, Patrick von Platen, Clara Ma, Yacine Jernite, Julien Plu, Canwen Xu, Teven Le~Scao, Sylvain Gugger, Mariama Drame, Quentin Lhoest, and Alexander Rush. 2020.
\newblock \href {https://doi.org/10.18653/v1/2020.emnlp-demos.6} {Transformers: State-of-the-art natural language processing}.
\newblock In \emph{Proceedings of the 2020 Conference on Empirical Methods in Natural Language Processing: System Demonstrations}, pages 38--45, Online. Association for Computational Linguistics.

\bibitem[{Yadav et~al.(2023)Yadav, Chandel, Chatufale, and Bandhakavi}]{yadav2023lahm}
Ankit Yadav, Shubham Chandel, Sushant Chatufale, and Anil Bandhakavi. 2023.
\newblock Lahm: Large annotated dataset for multi-domain and multilingual hate speech identification.
\newblock \emph{arXiv preprint arXiv:2304.00913}.

\bibitem[{Yin and Zubiaga(2021)}]{yin2021towards}
Wenjie Yin and Arkaitz Zubiaga. 2021.
\newblock Towards generalisable hate speech detection: a review on obstacles and solutions.
\newblock \emph{PeerJ Computer Science}, 7:e598.

\bibitem[{Yuan and Rizoiu(2022)}]{yuan2022detect}
Lanqin Yuan and Marian-Andrei Rizoiu. 2022.
\newblock Detect hate speech in unseen domains using multi-task learning: A case study of political public figures.
\newblock \emph{arXiv preprint arXiv:2208.10598}.

\bibitem[{Zhang and Luo(2019)}]{zhang2019hate}
Ziqi Zhang and Lei Luo. 2019.
\newblock Hate speech detection: A solved problem? the challenging case of long tail on twitter.
\newblock \emph{Semantic Web}, 10(5):925--945.

\end{thebibliography}

\section{Language Resource References}
\label{lr:ref}
\bibliographystylelanguageresource{lrec-coling2024-natbib}
\bibliographylanguageresource{lr}

\cleardoublepage
\appendix

\section{Data Release}
\label{section:datarelease}

It is very difficult to anonymise Twitter data to the extent that cannot be traced back from the text \cite{ayers_twitter_2018}, raising privacy concerns over the release of Twitter abuse datasets. While we recognise the prevalence of openly available Twitter hate speech datasets \cite{alkomah_twitter_2022}, due to institutional guidelines we are unable to release the annotated Tweets the make up the DoDo dataset, neither as anonymised text or as Tweet IDs only. We acknowledge that this limits reproducability, and we hope that the methodology outlined in \cref{section:add_dataset} demonstrates robustness and enables other researchers to emulate this study. We are able to make lists of accounts of public figures collated available to researchers on request, via emailing angusrwilliams@gmail.com.

\section{Data Annotation}
\label{section:add_annotation}

We used two different sets of annotators across the two domains, as we annotated the sets sequentially. Initial annotation rounds revealed high rates of annotator disagreement, with a large number of entries requiring expert annotation as a result. We use the same label schema for all domain and demographic pairs but use specific example tweets in the guidelines. We only employ annotators who pass a test of gold questions. Annotators were informed prior to accepting the task that the data would be used to train machine learning models as part of a research paper.

We employed 3,375 crowdworkers for male footballers and 3,513 for female footballers. Crowdworkers were paid \$0.20 per annotation, earning \$11.30/hour on average. Each entry was annotated by 3 crowdworkers, with an additional two annotations required if no majority agreement ($\frac{2}{3}$) was reached, then sent for expert annotation if still no majority agreement ($\frac{3}{5}$) was reached. The average annotator agreement per entry was 68\%, and the Cohen's kappa was 0.50.

For the MP datasets, we employed 23 high-quality annotators from a Trust \& Safety organisation. Annotators were paid \$0.33 per annotation, earning \$16.80/hour on average. Each entry received 3 annotations, then sent for expert annotation if no majority agreement was reached ($\frac{2}{3}$). The average entry-wise agreement was 82\% and the Cohen's kappa was 0.67.

\begin{figure*}[p]    
    \centering
    \captionsetup{justification=centering}
    \includegraphics[width=.93\textwidth]{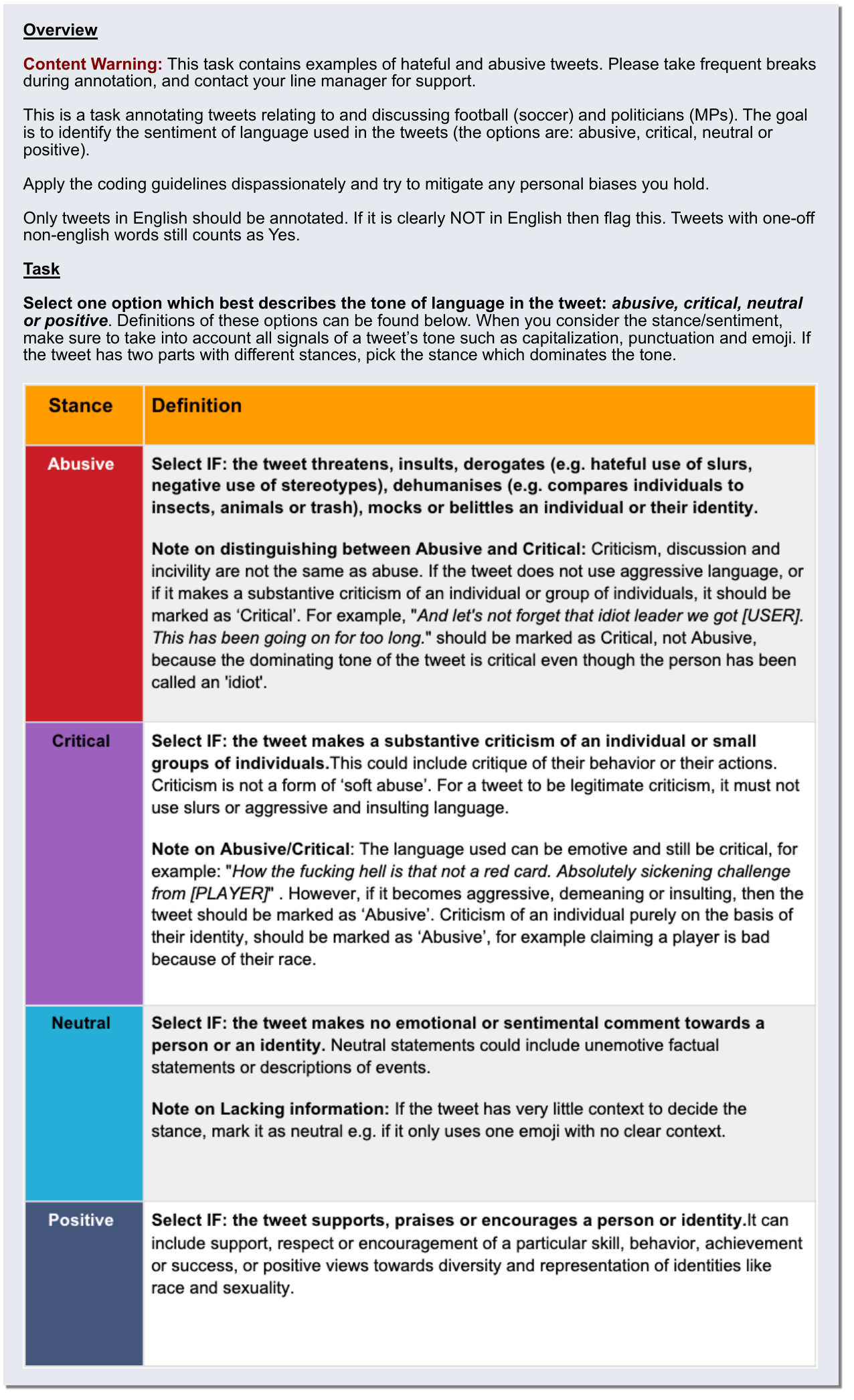}
    \caption{Instructions given to annotators.}
    \label{fig:annotinstructions}
\end{figure*}

\begin{figure*}[h]
    \centering
    \includegraphics[width=0.9\textwidth]{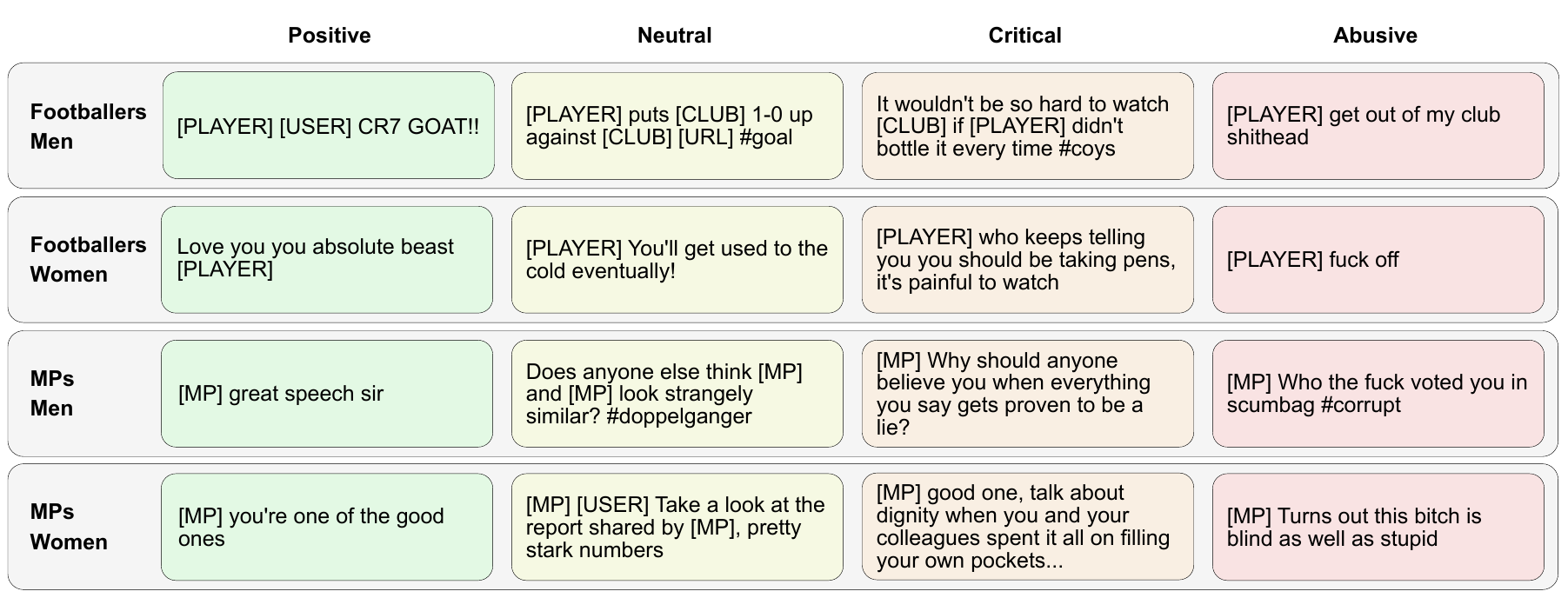}
    \caption{Fictional example tweets for each class label, loosely based on topics and sentiment of content in the dataset. Entries from the dataset are presented to annotators as shown, with special tokens to represent tagged mentions of public figures, accounts representing affiliations (e.g., football clubs), and other users. Examples are fictional as the dataset will not be released.}
    \label{fig:example_tweets}
\end{figure*}

An example of instructions given to annotators is displayed in \cref{fig:annotinstructions}. Fictional examples of tweet stances across domain-demographic pairs are visible in \cref{fig:example_tweets}. Due to the potentially harmful nature of the task, annotators were encouraged to regularly take breaks, and to contact their line manager in event of any problems or concerns. Annotator pay was above US minimum hourly wage on average.

\section{Data Statement}
\label{section:data_statement}

To document the generation and provenance of our dataset, we provide a data statement below \citep{bender_friedman_2018}.

\paragraph{Curation Rationale}

The purpose of the \textsc{DoDo} dataset is to train, evaluate, and refine language models for classification tasks related to understanding online conversations directed at footballers and MPs.

\paragraph{Language Variety}

Due to the UK-centric domains this dataset concerns (men's and women's UK football leagues, and UK MPs), all tweets are in English.

\paragraph{Speaker Demographics}

All entries are collected from Twitter and therefore generally represent the demographics of the platform. The sample is skewed towards those engaging in community discussion of the two domains on the platform (sports and politics).

\paragraph{Annotator Demographics}
The two domains used differing annotator pools. For the MPs data, we made use of a company offering annotation services that recruited 23 annotators to work for 5 weeks in early 2023. The annotators were screened from an initial pool of 36 annotators who took a test consisting of 36 difficult gold-standard questions (containing examples of all four class labels). The annotators had constant access to both a core team member from the service provider and from the core research team.

Fifteen annotators self-identified as women, and eight as men. The annotators were sent an optional survey to provide further information on their demographics. Out of 23 annotators, 21 responded to the survey. By age, 12 annotators were between 18-29 years old, eight were between 30-39 years old, and one was over 50 years old. In terms of completed education level, three annotators had high school degrees, eight annotators had undergraduate degrees, six annotators had postgraduate taught degrees, and four annotators had postgraduate research degrees. The majority of annotators were British (17), and other nationalities included Indian, Swedish, and United States. Twelve annotators identified as White, with one identifying as White Other and one identifying as White Arab. Other ethnicities included Black Caribbean (1), Indian (1), Indian British Asian (1), and Jewish (1). Most annotators identified as heterosexual (14), with other annotators identifying as bisexual (3), gay (1), and pansexual (1). Two chose not to disclose their sexuality. The majority stated that English was their native language (16), and four stated they were not native but fluent in the language. One chose not to disclose whether they were native English speakers or not. The majority of annotators disclosed that they spend 1-2 hours per day on social media (12). Four annotators stated that they spent, on average, less than 1 hour on social media per day (but more than 10 minutes), and five stated they spend more than 2 hours per day on social media. Some of the annotators reported having themselves been targeted by online abuse (9), with 11 reporting `never' and one preferring not to say.

The datasets for footballers were annotated separately using a crowdsourcing platform. Due to this, we have significantly less detail on the demographics of the users. The fb-m dataset was annotated by 3,375 crowdworkers from 41 countries. The fb-w dataset was annotated by 3,513 crowdworkers from 48 countries. The annotators for both datasets were primarily from Venezuela (56\% and 64\% respectively) and the United States (29\% and 18\% respectively).

\paragraph{Speech Situation}

The data consists of short-form written textual entries from social media (Twitter). These were presented and interpreted in isolation for labelling, i.e., not in a comment thread and without user/network or any additional information.

\paragraph{Text Characteristics}

The genre of texts is a mix of abusive, critical, positive, and neutral social media entries (tweets).

\section{Data Collection, Processing, and Sampling}
\label{section:add_dataset}

We chose to collect data on members of parliament and footballers: two types of well known public figure that both receive considerable amounts of online abuse but which operate in very different domains. These two domains also serve as useful bases because they have demographic diversity (in particular, they have both male and female participants, with gender being a well known source of difference in terms of abuse being received). 

We collect all tweets mentioning a public figure account, keeping only those that either directly reply to tweets written by public figures, or directly mention a public figure account without replying or referencing another tweet. We term these tweets \textit{audience contact}. From the audience contact tweets, we only consider tweets that contain some English text content aside from mentions and URLs. Where the Twitter API Filtered Stream endpoint did not return sufficient data for constructing an unlabelled pool, as was the case for female footballers, we made use of the Twitter API Full Archive Search endpoint to collect historic tweets. \cref{tab:pool_info} contains information on the unlabelled pools.

For each domain-demographic pair, starting with the unlabelled pool, we randomly sample (and remove) 3,000 entries for the test set and 1,000 entries for the validation set. We then randomly sample (and remove) 1,500 entries for training and concatenate these with a further 1,500 entries containing a keyword from a list of 731 abusive and hateful keywords (750 entries with at least one profanity keyword and 750 with at least one identity keyword), such that each training set has $3,000$ entries total. The list of keywords is compiled from \citet{davidsonAutomated2017, elsherief2018hate, vidgen2021learning, kirk2022hatemoji} and is available at \href{https://github.com/Turing-Online-Safety-Codebase/dodo-learning}{github.com/Turing-Online-Safety-Codebase/dodo-learning}. Each training set has 3,000 entries in total. \cref{table:strategy_counts} describes the counts of Tweets by stance for each sampling strategy used in the construction of datasets.

We replace all user mentions within tweets with tokens relating to the domain of the public figure mentioned before tweet annotation and use in training models. This does not completely anonymise tweets, as it does not account for other uses of names in tweet text.

\section{Additional Results}
\label{section:add_results}

\subsection{Where Unseen Performance Exceeds Seen Performance}
\label{section:add_unseen}
There are three cases where performance on unseen \texttt{dodos} exceeds performance on seen \texttt{dodos} in both full and fixed budget scenarios, visible in \cref{fig:dumbbells_single}. All three cases include \texttt{fb-m} in the training data, suggesting that the \texttt{fb-m} test set is more difficult that other \texttt{dodos}, or potentially that the \texttt{fb-m} training split is significantly different to the test split - further investigation is needed to fully understand this dynamic.

\subsection{Error Analysis}
\label{section:add_error_analysis}

Our error analysis is based on each fixed-budget single \texttt{dodo} model (i.e. \texttt{dodo1} experiments), evaluated on seen portions of the test set. We also analyse errors made by the fixed budget generalist model (i.e. \texttt{dodo4}), and shared errors made by all fixed budget condition models. We choose fixed budget models to ensure all models have seen the same total amount of training data. We present confusion matrices for all experiments in Fig. \ref{fig:confusion_grid}.

The fb-m model performed best on positive tweets (F1 = 0.86), and worst on critical tweets (F1 = 0.52). These results broadly hold for the fb-w model, which performed best on positive tweets (F1 = 0.91) and less well on abusive (F1 = 0.57) and critical (F1 = 0.52) tweets. The mp-m model performed best on critical tweets (F1 = 0.77), and worst on positive and neutral tweets (F1 = 0.69). As with footballers, these results broadly hold for the mp-w model, which performed best on critical tweets (F1 = 0.74), and less well on neutral (F1 = 0.66) and abusive tweets (F1 = 0.63). 

These results partly reflect class imbalance (the FBs data is heavily skewed towards positive tweets, the MPs data towards critical tweets), as well as some inherent similarity between classes which border one another i.e., positive vs. neutral, neutral vs. critical, and critical vs. abusive. Recurring errors reveal several tweet types that are challenging to classify: tweets that tweets that (i) contain a mixture of both positive and critical language; (ii) use positive or sarcastic language to mock; (iii) rely on emoji to convey abuse; (iv) contain niche insults; or (v) short, ambiguous tweets that lack context.

\subsection{Expanded Evaluation}

Here we provide expanded reference tables and figures on the results described in \cref{section:results}.

The per-class macro F1 score of each \texttt{dodo1} model and the two \texttt{dodo4} models evaluated on seen \texttt{dodos} are visible in \cref{table:f1_breakdown}, revealing relatively low performance on the critical and abusive classes for models trained on the two footballer datasets compared to the positive and neutral classes. For models trained on the MPs datasets, we see much less variation in per class performance.

\begin{figure*}[p]    
    \centering
    \captionsetup{justification=centering}
    \includegraphics[width=1\textwidth]{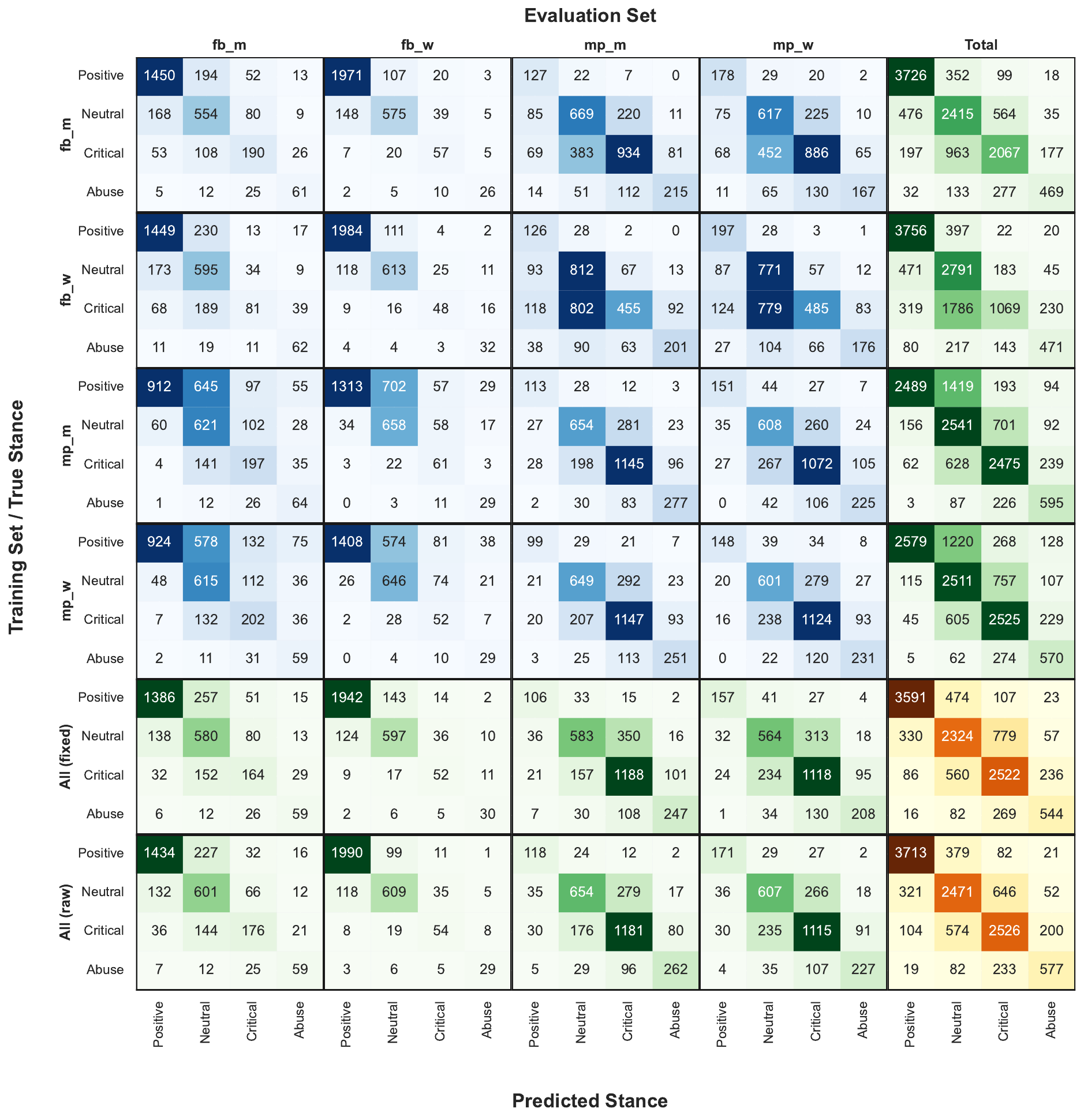}
    \caption{Grid of confusion matrices across chosen baselines, using soft voting across random seeds.}
    \label{fig:confusion_grid}
\end{figure*}

We also present a set of confusion matrices in \cref{fig:confusion_grid} for the specialist (\texttt{dodo1}), fixed budget generalist (\texttt{dodo4}, train size = 3,000), and full budget generalist (\texttt{dodo4}, train size = 12,000) models based on deBERT, evaluated on each evaluation set and the total evaluation set. 

Finally, we give a reference table of maximum Macro-F1 scores achieved by all baselines across all evaluation sets (\cref{tab:allf1}).

\begin{table}[H]
\footnotesize
\centering
\captionsetup{justification=centering}
\begin{tblr}{
  row{2} = {c},
  cell{1}{1} = {r=2}{},
  cell{1}{2} = {c=4}{c},
  cell{3}{2} = {c},
  cell{3}{3} = {c},
  cell{3}{4} = {c},
  cell{3}{5} = {c},
  cell{4}{2} = {c},
  cell{4}{3} = {c},
  cell{4}{4} = {c},
  cell{4}{5} = {c},
  cell{5}{2} = {c},
  cell{5}{3} = {c},
  cell{5}{4} = {c},
  cell{5}{5} = {c},
  cell{6}{2} = {c},
  cell{6}{3} = {c},
  cell{6}{4} = {c},
  cell{6}{5} = {c},
  cell{7}{2} = {c},
  cell{7}{3} = {c},
  cell{7}{4} = {c},
  cell{7}{5} = {c},
  cell{8}{2} = {c},
  cell{8}{3} = {c},
  cell{8}{4} = {c},
  cell{8}{5} = {c},
  hline{1,3,9} = {-}{},
  hline{2} = {2-5}{Gray},
}
\textbf{ dodo} & \textbf{Per-class F1 Scores} &                  &                   &                  \\
               & \textit{Positive}            & \textit{Neutral} & \textit{Critical} & \textit{Abusive} \\
fb\_m          & 0.86                         & 0.66             & 0.52              & 0.58             \\
fb\_w          & 0.94                         & 0.81             & 0.57              & 0.62             \\
mp\_m          & 0.69                         & 0.69             & 0.77              & 0.70             \\
mp\_w          & 0.72                         & 0.66             & 0.74              & 0.63             \\
All (fixed)    & 0.87                         & 0.67             & 0.71              & 0.61             \\
All (raw)      & 0.89                         & 0.71             & 0.73              & 0.66             
\end{tblr}
\caption{Per-class F1-scores for \texttt{dodo1} and \texttt{dodo4} baselines on seen evaluation sets.}
\label{table:f1_breakdown}
\end{table}

\begin{table*}[h]
    \centering
    \footnotesize
    \captionsetup{justification=centering}
    \begin{tabular}{@{\extracolsep{\fill}}lllllcc} 
    \hline
    \multirow{2}{*}{Domain}      & \multirow{2}{*}{Demographic} & \multirow{2}{*}{Pool Size} & \multicolumn{2}{c}{Collection Dates}                                  & \multicolumn{2}{c}{Collection Method}  \\ 
    \arrayrulecolor[rgb]{0.502,0.502,0.502}\cline{4-7}
                                 &                              &                            & \multicolumn{1}{c}{\textit{Start}} & \multicolumn{1}{c}{\textit{End}} & \textit{Streaming} & \textit{Search}   \\ 
    \arrayrulecolor{black}\hline
    \multirow{2}{*}{Footballers} & Men                          & 1,008,399                  & 12/08/2021                         & 02/02/2022                       & \checkmark                  &                   \\
                                 & Women                        & 226,689                    & 13/08/2021                         & 28/11/2022                       & \checkmark                  & \checkmark                 \\ 
    \arrayrulecolor[rgb]{0.502,0.502,0.502}\hline
    \multirow{2}{*}{MPs}         & Men                          & 1,000,000                  & 13/01/2022                         & 19/09/2022                       & \checkmark                  &                   \\
                                 & Women                        & 1,000,000                  & 13/01/2022                         & 19/09/2022                       & \checkmark                  &                   \\
    \arrayrulecolor{black}\hline
    \end{tabular}
    \caption{Dates and pool sizes for each domain-demographic pair.}
    \label{tab:pool_info}
\end{table*}

\begin{table*}[b]
\footnotesize
\centering
\captionsetup{justification=centering}
\begin{adjustbox}{width=1\textwidth}
\begin{tblr}{
  row{2} = {c},
  row{3} = {c},
  cell{1}{1} = {r=3}{},
  cell{1}{2} = {r=3}{},
  cell{1}{3} = {c=12}{c},
  cell{2}{3} = {c=4}{},
  cell{2}{7} = {c=4}{},
  cell{2}{11} = {c=4}{},
  cell{4}{1} = {r=4}{},
  cell{4}{3} = {r},
  cell{4}{4} = {r},
  cell{4}{5} = {r},
  cell{4}{6} = {r},
  cell{4}{7} = {r},
  cell{4}{8} = {r},
  cell{4}{9} = {r},
  cell{4}{10} = {r},
  cell{4}{11} = {r},
  cell{4}{12} = {r},
  cell{4}{13} = {r},
  cell{4}{14} = {r},
  cell{5}{3} = {r},
  cell{5}{4} = {r},
  cell{5}{5} = {r},
  cell{5}{6} = {r},
  cell{5}{7} = {r},
  cell{5}{8} = {r},
  cell{5}{9} = {r},
  cell{5}{10} = {r},
  cell{5}{11} = {r},
  cell{5}{12} = {r},
  cell{5}{13} = {r},
  cell{5}{14} = {r},
  cell{6}{3} = {r},
  cell{6}{4} = {r},
  cell{6}{5} = {r},
  cell{6}{6} = {r},
  cell{6}{7} = {r},
  cell{6}{8} = {r},
  cell{6}{9} = {r},
  cell{6}{10} = {r},
  cell{6}{11} = {r},
  cell{6}{12} = {r},
  cell{6}{13} = {r},
  cell{6}{14} = {r},
  cell{7}{3} = {r},
  cell{7}{4} = {r},
  cell{7}{5} = {r},
  cell{7}{6} = {r},
  cell{7}{7} = {r},
  cell{7}{8} = {r},
  cell{7}{9} = {r},
  cell{7}{10} = {r},
  cell{7}{11} = {r},
  cell{7}{12} = {r},
  cell{7}{13} = {r},
  cell{7}{14} = {r},
  cell{8}{1} = {r=4}{},
  cell{8}{3} = {r},
  cell{8}{4} = {r},
  cell{8}{5} = {r},
  cell{8}{6} = {r},
  cell{8}{7} = {r},
  cell{8}{8} = {r},
  cell{8}{9} = {r},
  cell{8}{10} = {r},
  cell{8}{11} = {r},
  cell{8}{12} = {r},
  cell{8}{13} = {r},
  cell{8}{14} = {r},
  cell{9}{3} = {r},
  cell{9}{4} = {r},
  cell{9}{5} = {r},
  cell{9}{6} = {r},
  cell{9}{7} = {r},
  cell{9}{8} = {r},
  cell{9}{9} = {r},
  cell{9}{10} = {r},
  cell{9}{11} = {r},
  cell{9}{12} = {r},
  cell{9}{13} = {r},
  cell{9}{14} = {r},
  cell{10}{3} = {r},
  cell{10}{4} = {r},
  cell{10}{5} = {r},
  cell{10}{6} = {r},
  cell{10}{7} = {r},
  cell{10}{8} = {r},
  cell{10}{9} = {r},
  cell{10}{10} = {r},
  cell{10}{11} = {r},
  cell{10}{12} = {r},
  cell{10}{13} = {r},
  cell{10}{14} = {r},
  cell{11}{3} = {r},
  cell{11}{4} = {r},
  cell{11}{5} = {r},
  cell{11}{6} = {r},
  cell{11}{7} = {r},
  cell{11}{8} = {r},
  cell{11}{9} = {r},
  cell{11}{10} = {r},
  cell{11}{11} = {r},
  cell{11}{12} = {r},
  cell{11}{13} = {r},
  cell{11}{14} = {r},
  cell{12}{1} = {r=4}{},
  cell{12}{3} = {r},
  cell{12}{4} = {r},
  cell{12}{5} = {r},
  cell{12}{6} = {r},
  cell{12}{7} = {r},
  cell{12}{8} = {r},
  cell{12}{9} = {r},
  cell{12}{10} = {r},
  cell{12}{11} = {r},
  cell{12}{12} = {r},
  cell{12}{13} = {r},
  cell{12}{14} = {r},
  cell{13}{3} = {r},
  cell{13}{4} = {r},
  cell{13}{5} = {r},
  cell{13}{6} = {r},
  cell{13}{7} = {r},
  cell{13}{8} = {r},
  cell{13}{9} = {r},
  cell{13}{10} = {r},
  cell{13}{11} = {r},
  cell{13}{12} = {r},
  cell{13}{13} = {r},
  cell{13}{14} = {r},
  cell{14}{3} = {r},
  cell{14}{4} = {r},
  cell{14}{5} = {r},
  cell{14}{6} = {r},
  cell{14}{7} = {r},
  cell{14}{8} = {r},
  cell{14}{9} = {r},
  cell{14}{10} = {r},
  cell{14}{11} = {r},
  cell{14}{12} = {r},
  cell{14}{13} = {r},
  cell{14}{14} = {r},
  cell{15}{3} = {r},
  cell{15}{4} = {r},
  cell{15}{5} = {r},
  cell{15}{6} = {r},
  cell{15}{7} = {r},
  cell{15}{8} = {r},
  cell{15}{9} = {r},
  cell{15}{10} = {r},
  cell{15}{11} = {r},
  cell{15}{12} = {r},
  cell{15}{13} = {r},
  cell{15}{14} = {r},
  cell{16}{1} = {r=4}{},
  cell{16}{3} = {r},
  cell{16}{4} = {r},
  cell{16}{5} = {r},
  cell{16}{6} = {r},
  cell{16}{7} = {r},
  cell{16}{8} = {r},
  cell{16}{9} = {r},
  cell{16}{10} = {r},
  cell{16}{11} = {r},
  cell{16}{12} = {r},
  cell{16}{13} = {r},
  cell{16}{14} = {r},
  cell{17}{3} = {r},
  cell{17}{4} = {r},
  cell{17}{5} = {r},
  cell{17}{6} = {r},
  cell{17}{7} = {r},
  cell{17}{8} = {r},
  cell{17}{9} = {r},
  cell{17}{10} = {r},
  cell{17}{11} = {r},
  cell{17}{12} = {r},
  cell{17}{13} = {r},
  cell{17}{14} = {r},
  cell{18}{3} = {r},
  cell{18}{4} = {r},
  cell{18}{5} = {r},
  cell{18}{6} = {r},
  cell{18}{7} = {r},
  cell{18}{8} = {r},
  cell{18}{9} = {r},
  cell{18}{10} = {r},
  cell{18}{11} = {r},
  cell{18}{12} = {r},
  cell{18}{13} = {r},
  cell{18}{14} = {r},
  cell{19}{3} = {r},
  cell{19}{4} = {r},
  cell{19}{5} = {r},
  cell{19}{6} = {r},
  cell{19}{7} = {r},
  cell{19}{8} = {r},
  cell{19}{9} = {r},
  cell{19}{10} = {r},
  cell{19}{11} = {r},
  cell{19}{12} = {r},
  cell{19}{13} = {r},
  cell{19}{14} = {r},
  vline{4,8} = {2}{Gray},
  vline{7,11} = {3-19}{Gray},
  hline{1,4,8,12,16,20} = {-}{},
  hline{2-3} = {3-14}{Gray},
}
\textbf{Split} & \texttt{dodo}  & \textbf{Sampling Strategy} &                   &                  &                   &                    &                   &                  &                   &                   &                   &                  &                   \\
               &                & Random                     &                   &                  &                   & Profanity Keywords &                   &                  &                   & Identity Keywords &                   &                  &                   \\
               &                & \textit{Abusive}           & \textit{Critical} & \textit{Neutral} & \textit{Positive} & \textit{Abusive}   & \textit{Critical} & \textit{Neutral} & \textit{Positive} & \textit{Abusive}  & \textit{Critical} & \textit{Neutral} & \textit{Positive} \\
Train          & fb\_m          & 45                         & 172               & 531              & 752               & 290                & 224               & 52               & 184               & 532               & 79                & 64               & 75                \\
               & fb\_w          & 18                         & 63                & 432              & 987               & 346                & 190               & 211              & 467               & 117               & 29                & 76               & 64                \\
               & mp\_m          & 212                        & 725               & 471              & 92                & 372                & 311               & 57               & 10                & 423               & 247               & 77               & 3                 \\
               & mp\_w          & 153                        & 746               & 477              & 124               & 349                & 322               & 67               & 12                & 368               & 285               & 84               & 13                \\
Test           & fb\_m          & 103                        & 377               & 811              & 1709              & 0                  & 0                 & 0                & 0                 & 0                 & 0                 & 0                & 0                 \\
               & fb\_w          & 43                         & 89                & 767              & 2101              & 0                  & 0                 & 0                & 0                 & 0                 & 0                 & 0                & 0                 \\
               & mp\_m          & 392                        & 1467              & 985              & 156               & 0                  & 0                 & 0                & 0                 & 0                 & 0                 & 0                & 0                 \\
               & mp\_w          & 373                        & 1471              & 927              & 229               & 0                  & 0                 & 0                & 0                 & 0                 & 0                 & 0                & 0                 \\
Validation     & fb\_m          & 33                         & 93                & 335              & 539               & 0                  & 0                 & 0                & 0                 & 0                 & 0                 & 0                & 0                 \\
               & fb\_w          & 14                         & 45                & 267              & 674               & 0                  & 0                 & 0                & 0                 & 0                 & 0                 & 0                & 0                 \\
               & mp\_m          & 140                        & 484               & 332              & 44                & 0                  & 0                 & 0                & 0                 & 0                 & 0                 & 0                & 0                 \\
               & mp\_w          & 135                        & 459               & 337              & 69                & 0                  & 0                 & 0                & 0                 & 0                 & 0                 & 0                & 0                 \\
\textbf{Total} & \textbf{fb\_m} & \textbf{181}               & \textbf{642}      & \textbf{1677}    & \textbf{3000}     & \textbf{290}       & \textbf{224}      & \textbf{52}      & \textbf{184}      & \textbf{532}      & \textbf{79}       & \textbf{64}      & \textbf{75}       \\
               & \textbf{fb\_w} & \textbf{75}                & \textbf{197}      & \textbf{1466}    & \textbf{3762}     & \textbf{346}       & \textbf{190}      & \textbf{211}     & \textbf{467}      & \textbf{117}      & \textbf{29}       & \textbf{76}      & \textbf{64}       \\
               & \textbf{mp\_m} & \textbf{744}               & \textbf{2676}     & \textbf{1788}    & \textbf{292}      & \textbf{372}       & \textbf{311}      & \textbf{57}      & \textbf{10}       & \textbf{423}      & \textbf{247}      & \textbf{77}      & \textbf{3}        \\
               & \textbf{mp\_w} & \textbf{661}               & \textbf{2676}     & \textbf{1741}    & \textbf{422}      & \textbf{349}       & \textbf{322}      & \textbf{67}      & \textbf{12}       & \textbf{368}      & \textbf{285}      & \textbf{84}      & \textbf{13}       
\end{tblr}
\end{adjustbox}
\caption{Tweet counts for \texttt{dodo} splits across sampling strategy and stance.}
\label{table:strategy_counts}
\end{table*}

\begin{table*}[tb]
\centering
\footnotesize
\begin{tabular}{lcccc|cc|ccccc} 
\toprule
\multicolumn{1}{c}{}              & \multicolumn{4}{c}{\textbf{Train On }} &                         &              & \multicolumn{5}{c}{\textbf{Test On }}                                                                                                                                                                                      \\ 
\cline{2-12}
                                  & fb-m  & fb-w  & mp-m  & mp-w           & model                   & budget       & total                                     & fb-m                                      & fb-w                                      & mp-m                                      & mp-w                                       \\ 
\hhline{~-----------}
\multirow{8}{*}{\textbf{dodo1 }}  & \checkmark &       &       &                & deBERT                  & fixed = full & {\cellcolor[rgb]{0.753,0.902,0.827}}0.688 & {\cellcolor[rgb]{0.882,0.953,0.918}}0.656 & {\cellcolor[rgb]{0.627,0.851,0.741}}0.719 & {\cellcolor[rgb]{0.973,0.992,0.984}}0.633 & {\cellcolor[rgb]{0.992,0.957,0.953}}0.609  \\
                                  & \checkmark &       &       &                & diBERT                  & fixed = full & {\cellcolor[rgb]{0.984,0.937,0.929}}0.600   & {\cellcolor[rgb]{0.976,0.886,0.878}}0.580  & {\cellcolor[rgb]{0.98,0.91,0.902}}0.589   & {\cellcolor[rgb]{0.949,0.737,0.722}}0.518 & {\cellcolor[rgb]{0.949,0.749,0.729}}0.522  \\ 
\hhline{~-----------}
                                  &       & \checkmark &       &                & deBERT                  & fixed = full & {\cellcolor[rgb]{0.992,1,0.996}}0.628     & {\cellcolor[rgb]{0.98,0.902,0.894}}0.586  & {\cellcolor[rgb]{0.8,0.922,0.863}}0.676   & {\cellcolor[rgb]{0.957,0.788,0.773}}0.539 & {\cellcolor[rgb]{0.961,0.804,0.788}}0.545  \\
                                  &       & \checkmark &       &                & diBERT                  & fixed = full & {\cellcolor[rgb]{0.945,0.714,0.694}}0.508 & {\cellcolor[rgb]{0.929,0.635,0.612}}0.476 & {\cellcolor[rgb]{0.992,0.973,0.969}}0.615 & {\cellcolor[rgb]{0.902,0.49,0.455}}0.415  & {\cellcolor[rgb]{0.902,0.486,0.451}}0.413  \\ 
\hhline{~-----------}
                                  &       &       & \checkmark &                & deBERT                  & fixed = full & {\cellcolor[rgb]{0.847,0.937,0.894}}0.665 & {\cellcolor[rgb]{0.957,0.78,0.765}}0.536  & {\cellcolor[rgb]{0.976,0.878,0.871}}0.576 & {\cellcolor[rgb]{0.663,0.867,0.769}}0.71  & {\cellcolor[rgb]{0.847,0.937,0.894}}0.665  \\
                                  &       &       & \checkmark &                & diBERT                  & fixed = full & {\cellcolor[rgb]{0.973,0.867,0.855}}0.571 & {\cellcolor[rgb]{0.91,0.545,0.514}}0.438  & {\cellcolor[rgb]{0.91,0.541,0.51}}0.437   & {\cellcolor[rgb]{0.996,0.98,0.98}}0.619   & {\cellcolor[rgb]{0.98,0.906,0.898}}0.587   \\ 
\hhline{~-----------}
                                  &       &       &       & \checkmark          & deBERT                  & fixed = full & {\cellcolor[rgb]{0.804,0.922,0.867}}0.675 & {\cellcolor[rgb]{0.961,0.812,0.8}}0.549   & {\cellcolor[rgb]{0.976,0.882,0.875}}0.578 & {\cellcolor[rgb]{0.78,0.914,0.847}}0.681  & {\cellcolor[rgb]{0.773,0.91,0.843}}0.683   \\
                                  &       &       &       & \checkmark          & diBERT                  & fixed = full & {\cellcolor[rgb]{0.98,0.898,0.89}}0.584   & {\cellcolor[rgb]{0.918,0.573,0.541}}0.449 & {\cellcolor[rgb]{0.914,0.565,0.533}}0.446 & {\cellcolor[rgb]{0.984,0.918,0.91}}0.592  & {\cellcolor[rgb]{0.988,0.949,0.945}}0.605  \\ 
\hline
\multirow{24}{*}{\textbf{dodo2 }} & \checkmark & \checkmark &       &                & \multirow{2}{*}{deBERT} & fixed        & {\cellcolor[rgb]{0.831,0.933,0.886}}0.668 & {\cellcolor[rgb]{0.957,0.984,0.973}}0.637 & {\cellcolor[rgb]{0.341,0.733,0.541}}\textbf{0.790*}  & {\cellcolor[rgb]{0.98,0.906,0.902}}0.588  & {\cellcolor[rgb]{0.976,0.886,0.878}}0.579  \\
                                  & \checkmark & \checkmark &       &                &                         & full         & {\cellcolor[rgb]{0.831,0.933,0.886}}0.668 & {\cellcolor[rgb]{0.949,0.98,0.965}}0.639  & {\cellcolor[rgb]{0.667,0.867,0.769}}0.709 & {\cellcolor[rgb]{0.984,0.925,0.922}}0.596 & {\cellcolor[rgb]{0.984,0.922,0.914}}0.594  \\
                                  & \checkmark & \checkmark &       &                & \multirow{2}{*}{diBERT} & fixed        & {\cellcolor[rgb]{0.976,0.878,0.871}}0.577 & {\cellcolor[rgb]{0.965,0.831,0.82}}0.557  & {\cellcolor[rgb]{0.984,0.918,0.914}}0.593 & {\cellcolor[rgb]{0.937,0.678,0.659}}0.494 & {\cellcolor[rgb]{0.941,0.698,0.675}}0.501  \\
                                  & \checkmark & \checkmark &       &                &                         & full         & {\cellcolor[rgb]{0.992,0.961,0.961}}0.611 & {\cellcolor[rgb]{0.98,0.902,0.894}}0.586  & {\cellcolor[rgb]{0.992,0.961,0.957}}0.61  & {\cellcolor[rgb]{0.949,0.745,0.725}}0.521 & {\cellcolor[rgb]{0.949,0.741,0.722}}0.519  \\ 
\hhline{~-----------}
                                  & \checkmark &       & \checkmark &                & \multirow{2}{*}{deBERT} & fixed        & {\cellcolor[rgb]{0.651,0.859,0.757}}0.713 & {\cellcolor[rgb]{0.969,0.988,0.98}}0.634  & {\cellcolor[rgb]{0.616,0.847,0.733}}0.722 & {\cellcolor[rgb]{0.761,0.906,0.835}}0.686 & {\cellcolor[rgb]{0.878,0.953,0.914}}0.657  \\
                                  & \checkmark &       & \checkmark &                &                         & full         & {\cellcolor[rgb]{0.608,0.843,0.729}}0.724 & {\cellcolor[rgb]{0.871,0.949,0.91}}0.659  & {\cellcolor[rgb]{0.686,0.875,0.78}}0.705  & {\cellcolor[rgb]{0.69,0.875,0.784}}0.704  & {\cellcolor[rgb]{0.827,0.933,0.882}}0.669  \\
                                  & \checkmark &       & \checkmark &                & \multirow{2}{*}{diBERT} & fixed        & {\cellcolor[rgb]{0.898,0.961,0.929}}0.652 & {\cellcolor[rgb]{0.973,0.859,0.847}}0.568 & {\cellcolor[rgb]{0.98,0.906,0.902}}0.588  & {\cellcolor[rgb]{0.988,0.941,0.937}}0.602 & {\cellcolor[rgb]{0.984,0.922,0.914}}0.594  \\
                                  & \checkmark &       & \checkmark &                &                         & full         & {\cellcolor[rgb]{0.82,0.929,0.875}}0.671  & {\cellcolor[rgb]{0.984,0.929,0.925}}0.598 & {\cellcolor[rgb]{0.988,0.953,0.953}}0.608 & {\cellcolor[rgb]{0.992,0.969,0.965}}0.613 & {\cellcolor[rgb]{0.992,0.961,0.957}}0.61   \\ 
\hhline{~-----------}
                                  & \checkmark &       &       & \checkmark          & \multirow{2}{*}{deBERT} & fixed        & {\cellcolor[rgb]{0.643,0.859,0.753}}0.715 & {\cellcolor[rgb]{0.922,0.969,0.945}}0.646 & {\cellcolor[rgb]{0.847,0.937,0.894}}0.665 & {\cellcolor[rgb]{0.741,0.898,0.82}}0.691  & {\cellcolor[rgb]{0.82,0.929,0.875}}0.671   \\
                                  & \checkmark &       &       & \checkmark          &                         & full         & {\cellcolor[rgb]{0.608,0.843,0.729}}0.724 & {\cellcolor[rgb]{0.875,0.949,0.914}}0.658 & {\cellcolor[rgb]{0.745,0.898,0.824}}0.69  & {\cellcolor[rgb]{0.729,0.89,0.812}}0.694  & {\cellcolor[rgb]{0.78,0.914,0.847}}0.681   \\
                                  & \checkmark &       &       & \checkmark          & \multirow{2}{*}{diBERT} & fixed        & {\cellcolor[rgb]{0.918,0.969,0.945}}0.647 & {\cellcolor[rgb]{0.969,0.847,0.839}}0.564 & {\cellcolor[rgb]{0.98,0.906,0.898}}0.587  & {\cellcolor[rgb]{0.976,0.886,0.878}}0.58  & {\cellcolor[rgb]{0.984,0.922,0.918}}0.595  \\
                                  & \checkmark &       &       & \checkmark          &                         & full         & {\cellcolor[rgb]{0.847,0.937,0.894}}0.665 & {\cellcolor[rgb]{0.98,0.91,0.906}}0.59    & {\cellcolor[rgb]{0.984,0.922,0.914}}0.594 & {\cellcolor[rgb]{0.992,0.961,0.961}}0.611 & {\cellcolor[rgb]{0.992,0.969,0.965}}0.613  \\ 
\hhline{~-----------}
                                  &       & \checkmark & \checkmark &                & \multirow{2}{*}{deBERT} & fixed        & {\cellcolor[rgb]{0.694,0.878,0.788}}0.703 & {\cellcolor[rgb]{0.988,0.949,0.945}}0.606 & {\cellcolor[rgb]{0.729,0.89,0.812}}0.694  & {\cellcolor[rgb]{0.82,0.929,0.875}}0.671  & {\cellcolor[rgb]{0.922,0.969,0.945}}0.646  \\
                                  &       & \checkmark & \checkmark &                &                         & full         & {\cellcolor[rgb]{0.62,0.847,0.737}}0.721  & {\cellcolor[rgb]{0.988,0.953,0.953}}0.608 & {\cellcolor[rgb]{0.71,0.882,0.796}}0.699  & {\cellcolor[rgb]{0.663,0.867,0.769}}0.71  & {\cellcolor[rgb]{0.827,0.933,0.882}}0.669  \\
                                  &       & \checkmark & \checkmark &                & \multirow{2}{*}{diBERT} & fixed        & {\cellcolor[rgb]{0.918,0.969,0.945}}0.647 & {\cellcolor[rgb]{0.937,0.678,0.659}}0.494 & {\cellcolor[rgb]{0.992,0.973,0.969}}0.615 & {\cellcolor[rgb]{0.976,0.89,0.882}}0.581  & {\cellcolor[rgb]{0.976,0.875,0.867}}0.575  \\
                                  &       & \checkmark & \checkmark &                &                         & full         & {\cellcolor[rgb]{0.949,0.98,0.965}}0.639  & {\cellcolor[rgb]{0.937,0.686,0.663}}0.496 & {\cellcolor[rgb]{0.976,0.875,0.867}}0.575 & {\cellcolor[rgb]{0.988,0.945,0.941}}0.604 & {\cellcolor[rgb]{0.98,0.91,0.902}}0.589    \\ 
\hhline{~-----------}
                                  &       & \checkmark &       & \checkmark          & \multirow{2}{*}{deBERT} & fixed        & {\cellcolor[rgb]{0.675,0.871,0.773}}0.708 & {\cellcolor[rgb]{0.988,0.945,0.941}}0.604 & {\cellcolor[rgb]{0.788,0.918,0.855}}0.679 & {\cellcolor[rgb]{0.867,0.945,0.906}}0.66  & {\cellcolor[rgb]{0.835,0.933,0.886}}0.667  \\
                                  &       & \checkmark &       & \checkmark          &                         & full         & {\cellcolor[rgb]{0.616,0.847,0.733}}0.722 & {\cellcolor[rgb]{0.992,0.965,0.961}}0.612 & {\cellcolor[rgb]{0.757,0.902,0.831}}0.687 & {\cellcolor[rgb]{0.725,0.89,0.808}}0.695  & {\cellcolor[rgb]{0.769,0.906,0.839}}0.684  \\
                                  &       & \checkmark &       & \checkmark          & \multirow{2}{*}{diBERT} & fixed        & {\cellcolor[rgb]{0.988,0.996,0.992}}0.629 & {\cellcolor[rgb]{0.945,0.722,0.706}}0.512 & {\cellcolor[rgb]{0.973,0.859,0.851}}0.569 & {\cellcolor[rgb]{0.973,0.855,0.847}}0.567 & {\cellcolor[rgb]{0.973,0.867,0.855}}0.571  \\
                                  &       & \checkmark &       & \checkmark          &                         & full         & {\cellcolor[rgb]{0.953,0.984,0.969}}0.638 & {\cellcolor[rgb]{0.945,0.722,0.702}}0.511 & {\cellcolor[rgb]{0.976,0.875,0.867}}0.575 & {\cellcolor[rgb]{0.98,0.914,0.906}}0.591  & {\cellcolor[rgb]{0.992,0.961,0.961}}0.611  \\ 
\hhline{~-----------}
                                  &       &       & \checkmark & \checkmark          & \multirow{2}{*}{deBERT} & fixed        & {\cellcolor[rgb]{0.851,0.941,0.894}}0.664 & {\cellcolor[rgb]{0.957,0.773,0.757}}0.533 & {\cellcolor[rgb]{0.965,0.827,0.816}}0.556 & {\cellcolor[rgb]{0.816,0.925,0.875}}0.672 & {\cellcolor[rgb]{0.773,0.91,0.843}}0.683   \\
                                  &       &       & \checkmark & \checkmark          &                         & full         & {\cellcolor[rgb]{0.773,0.91,0.843}}0.683  & {\cellcolor[rgb]{0.969,0.835,0.824}}0.559 & {\cellcolor[rgb]{0.976,0.875,0.867}}0.575 & {\cellcolor[rgb]{0.737,0.894,0.816}}0.692 & {\cellcolor[rgb]{0.757,0.902,0.831}}0.687  \\
                                  &       &       & \checkmark & \checkmark          & \multirow{2}{*}{diBERT} & fixed        & {\cellcolor[rgb]{0.973,0.875,0.863}}0.574 & {\cellcolor[rgb]{0.918,0.584,0.553}}0.454 & {\cellcolor[rgb]{0.902,0.49,0.455}}0.416  & {\cellcolor[rgb]{0.992,0.957,0.953}}0.609 & {\cellcolor[rgb]{0.984,0.929,0.925}}0.598  \\
                                  &       &       & \checkmark & \checkmark          &                         & full         & {\cellcolor[rgb]{0.996,0.992,0.992}}0.624 & {\cellcolor[rgb]{0.937,0.675,0.651}}0.492 & {\cellcolor[rgb]{0.941,0.69,0.671}}0.499  & {\cellcolor[rgb]{0.969,0.988,0.98}}0.634  & {\cellcolor[rgb]{0.984,0.996,0.992}}0.63   \\ 
\hline
\multirow{16}{*}{\textbf{dodo3 }} & \checkmark & \checkmark & \checkmark &                & \multirow{2}{*}{deBERT} & fixed        & {\cellcolor[rgb]{0.663,0.867,0.769}}0.71  & {\cellcolor[rgb]{0.988,0.996,0.992}}0.629 & {\cellcolor[rgb]{0.557,0.82,0.69}}0.737   & {\cellcolor[rgb]{0.824,0.929,0.878}}0.67  & {\cellcolor[rgb]{0.91,0.965,0.937}}0.649   \\
                                  & \checkmark & \checkmark & \checkmark &                &                         & full         & {\cellcolor[rgb]{0.62,0.847,0.737}}0.721  & {\cellcolor[rgb]{0.996,0.992,0.992}}0.623 & {\cellcolor[rgb]{0.561,0.824,0.694}}0.736 & {\cellcolor[rgb]{0.702,0.878,0.792}}0.701 & {\cellcolor[rgb]{0.851,0.941,0.894}}0.664  \\
                                  & \checkmark & \checkmark & \checkmark &                & \multirow{2}{*}{diBERT} & fixed        & {\cellcolor[rgb]{0.961,0.984,0.973}}0.636 & {\cellcolor[rgb]{0.965,0.82,0.808}}0.552  & {\cellcolor[rgb]{0.984,0.929,0.925}}0.598 & {\cellcolor[rgb]{0.976,0.878,0.871}}0.576 & {\cellcolor[rgb]{0.969,0.851,0.839}}0.565  \\
                                  & \checkmark & \checkmark & \checkmark &                &                         & full         & {\cellcolor[rgb]{0.871,0.949,0.91}}0.659  & {\cellcolor[rgb]{0.976,0.878,0.871}}0.577 & {\cellcolor[rgb]{0.992,0.961,0.961}}0.611 & {\cellcolor[rgb]{0.992,0.973,0.973}}0.616 & {\cellcolor[rgb]{0.98,0.914,0.906}}0.591   \\ 
\hhline{~-----------}
                                  & \checkmark & \checkmark &       & \checkmark          & \multirow{2}{*}{deBERT} & fixed        & {\cellcolor[rgb]{0.714,0.886,0.8}}0.698   & {\cellcolor[rgb]{0.992,0.969,0.969}}0.614 & {\cellcolor[rgb]{0.612,0.843,0.729}}0.723 & {\cellcolor[rgb]{0.965,0.988,0.976}}0.635 & {\cellcolor[rgb]{0.961,0.984,0.973}}0.636  \\
                                  & \checkmark & \checkmark &       & \checkmark          &                         & full         & {\cellcolor[rgb]{0.569,0.827,0.698}}0.734 & {\cellcolor[rgb]{0.914,0.965,0.941}}0.648 & {\cellcolor[rgb]{0.6,0.839,0.722}}0.726   & {\cellcolor[rgb]{0.729,0.89,0.812}}0.694  & {\cellcolor[rgb]{0.776,0.91,0.847}}0.682   \\
                                  & \checkmark & \checkmark &       & \checkmark          & \multirow{2}{*}{diBERT} & fixed        & {\cellcolor[rgb]{0.996,0.996,0.996}}0.625 & {\cellcolor[rgb]{0.957,0.776,0.761}}0.534 & {\cellcolor[rgb]{0.976,0.878,0.871}}0.576 & {\cellcolor[rgb]{0.965,0.824,0.812}}0.553 & {\cellcolor[rgb]{0.965,0.816,0.804}}0.55   \\
                                  & \checkmark & \checkmark &       & \checkmark          &                         & full         & {\cellcolor[rgb]{0.816,0.925,0.875}}0.672 & {\cellcolor[rgb]{0.976,0.878,0.871}}0.576 & {\cellcolor[rgb]{0.969,0.988,0.98}}0.634  & {\cellcolor[rgb]{0.98,0.914,0.906}}0.591  & {\cellcolor[rgb]{0.988,0.949,0.945}}0.605  \\ 
\hhline{~-----------}
                                  & \checkmark &       & \checkmark & \checkmark          & \multirow{2}{*}{deBERT} & fixed        & {\cellcolor[rgb]{0.651,0.859,0.757}}0.713 & 0.626                                     & {\cellcolor[rgb]{0.82,0.929,0.875}}0.671  & {\cellcolor[rgb]{0.765,0.906,0.835}}0.685 & {\cellcolor[rgb]{0.812,0.925,0.871}}0.673  \\
                                  & \checkmark &       & \checkmark & \checkmark          &                         & full         & {\cellcolor[rgb]{0.561,0.824,0.694}}\textbf{0.736*} & {\cellcolor[rgb]{0.851,0.941,0.894}}\textbf{0.664*} & {\cellcolor[rgb]{0.682,0.871,0.776}}0.706 & {\cellcolor[rgb]{0.655,0.863,0.761}}0.712 & {\cellcolor[rgb]{0.737,0.894,0.816}}\textbf{0.692*}  \\
                                  & \checkmark &       & \checkmark & \checkmark          & \multirow{2}{*}{diBERT} & fixed        & {\cellcolor[rgb]{0.914,0.965,0.941}}0.648 & {\cellcolor[rgb]{0.965,0.831,0.82}}0.557  & {\cellcolor[rgb]{0.98,0.906,0.898}}0.587  & {\cellcolor[rgb]{0.988,0.941,0.937}}0.602 & {\cellcolor[rgb]{0.992,0.957,0.953}}0.609  \\
                                  & \checkmark &       & \checkmark & \checkmark          &                         & full         & {\cellcolor[rgb]{0.808,0.925,0.867}}0.674 & {\cellcolor[rgb]{0.976,0.894,0.886}}0.583 & {\cellcolor[rgb]{0.984,0.918,0.914}}0.593 & {\cellcolor[rgb]{0.973,0.992,0.984}}0.633 & 0.626                                      \\ 
\hhline{~-----------}
                                  &       & \checkmark & \checkmark & \checkmark          & \multirow{2}{*}{deBERT} & fixed        & {\cellcolor[rgb]{0.725,0.89,0.808}}0.695  & {\cellcolor[rgb]{0.98,0.898,0.894}}0.585  & {\cellcolor[rgb]{0.855,0.941,0.898}}0.663 & {\cellcolor[rgb]{0.894,0.957,0.925}}0.653 & {\cellcolor[rgb]{0.875,0.949,0.914}}0.658  \\
                                  &       & \checkmark & \checkmark & \checkmark          &                         & full         & {\cellcolor[rgb]{0.608,0.843,0.729}}0.724 & {\cellcolor[rgb]{0.98,0.914,0.906}}0.591  & {\cellcolor[rgb]{0.729,0.89,0.812}}0.694  & {\cellcolor[rgb]{0.639,0.855,0.749}}\textbf{0.716*} & {\cellcolor[rgb]{0.737,0.894,0.816}}\textbf{0.692*}  \\
                                  &       & \checkmark & \checkmark & \checkmark          & \multirow{2}{*}{diBERT} & fixed        & {\cellcolor[rgb]{0.937,0.976,0.957}}0.642 & {\cellcolor[rgb]{0.933,0.667,0.643}}0.488 & {\cellcolor[rgb]{0.973,0.859,0.851}}0.569 & {\cellcolor[rgb]{0.984,0.918,0.91}}0.592  & {\cellcolor[rgb]{0.988,0.941,0.937}}0.602  \\
                                  &       & \checkmark & \checkmark & \checkmark          &                         & full         & {\cellcolor[rgb]{0.855,0.941,0.898}}0.663 & {\cellcolor[rgb]{0.949,0.733,0.714}}0.516 & {\cellcolor[rgb]{0.98,0.902,0.894}}0.586  & {\cellcolor[rgb]{0.992,0.969,0.969}}0.614 & {\cellcolor[rgb]{0.996,0.98,0.976}}0.618   \\ 
\hline
\multirow{4}{*}{\textbf{dodo4 }}  & \checkmark & \checkmark & \checkmark & \checkmark          & \multirow{2}{*}{deBERT} & fixed        & {\cellcolor[rgb]{0.678,0.871,0.776}}0.707 & {\cellcolor[rgb]{0.945,0.98,0.965}}0.64   & {\cellcolor[rgb]{0.694,0.878,0.788}}0.703 & {\cellcolor[rgb]{0.855,0.941,0.898}}0.663 & {\cellcolor[rgb]{0.89,0.957,0.925}}0.654   \\
                                  & \checkmark & \checkmark & \checkmark & \checkmark          &                         & full         & {\cellcolor[rgb]{0.592,0.835,0.718}}0.728 & {\cellcolor[rgb]{0.969,0.988,0.98}}0.634  & {\cellcolor[rgb]{0.651,0.859,0.757}}0.713 & {\cellcolor[rgb]{0.667,0.867,0.769}}0.709 & {\cellcolor[rgb]{0.769,0.906,0.839}}0.684  \\
                                  & \checkmark & \checkmark & \checkmark & \checkmark          & \multirow{2}{*}{diBERT} & fixed        & {\cellcolor[rgb]{0.929,0.973,0.953}} 0.644 & {\cellcolor[rgb]{0.957,0.773,0.757}}0.533 & {\cellcolor[rgb]{0.98,0.914,0.906}}0.591  & {\cellcolor[rgb]{0.976,0.886,0.878}}0.58  & {\cellcolor[rgb]{0.976,0.886,0.878}}0.579  \\
                                  & \checkmark & \checkmark & \checkmark & \checkmark          &                         & full         & {\cellcolor[rgb]{0.765,0.906,0.835}}0.685 & {\cellcolor[rgb]{0.98,0.91,0.902}}0.589   & {\cellcolor[rgb]{0.949,0.98,0.965}}0.639  & {\cellcolor[rgb]{0.973,0.992,0.984}}0.633 & {\cellcolor[rgb]{0.973,0.992,0.984}}0.633  \\
\bottomrule
\end{tabular}
\caption{Macro-F1 score for all sets of baseline models (maximum value across three seeds). Best Macro-F1 per test set (total and each of the four \texttt{dodo} splits) is bold and starred. Colour-coded according to increasing Macro-F1 Score.}
\label{tab:allf1}
\end{table*}

\end{document}